\documentclass{article}

    \usepackage[final, nonatbib]{neurips_2022}

\usepackage[utf8]{inputenc} 
\usepackage[T1]{fontenc}    
\usepackage{hyperref}       
\usepackage{url}            
\usepackage{booktabs}       
\usepackage{amsfonts}       
\usepackage{nicefrac}       
\usepackage{microtype}      
\usepackage{xcolor}         
\usepackage{subfig}
\usepackage{graphicx}
\usepackage{graphicx}
\usepackage{amsmath}         
\usepackage{csquotes}
\usepackage{tikz} 
\usetikzlibrary{calc} 
\tikzstyle{vertex} = [fill,shape=circle,node distance=10pt]
\tikzstyle{edge} = [fill,opacity=.5,fill opacity=.5,line cap=round, line join=round, line width=30pt]
\tikzstyle{elabel} =  [fill,shape=circle,node distance=20pt]
\pgfdeclarelayer{background}
\pgfsetlayers{background,main}
\usetikzlibrary{backgrounds,fit}
\tikzstyle{block} = [draw, fill=yellow!40, rectangle, minimum height=3em, minimum width=4em]
\tikzstyle{sum} = [draw, fill=white, circle, scale=0.005, node distance=0.5cm]
\tikzstyle{input} = [coordinate]
\tikzstyle{output} = [coordinate]  
\tikzstyle{pinstyle} = [pin edge={to-,thin,black}]
\usetikzlibrary{shapes,arrows}
\usepackage{adjustbox}
\usepackage{multirow}
\usepackage{bbm}
\usetikzlibrary{shapes,arrows}
\usepackage{adjustbox}
\tikzstyle{block} = [draw, fill=yellow!50, rectangle, minimum height=3em, minimum width=4em]
\tikzstyle{sum} = [draw, fill=white, circle, scale=0.005, node distance=0.5cm]
\tikzstyle{input} = [coordinate]
\tikzstyle{output} = [coordinate]  
\tikzstyle{pinstyle} = [pin edge={to-,thin,black}]
\usepackage{csquotes}

\usepackage{algorithm}
\usepackage{algpseudocode}

\title{Hypergraph Learning based Recommender System for Anomaly Detection, Control and Optimization}

\author{
  Sakhinana Sagar Srinivas\thanks{Lead Author}, \quad Rajat Kumar Sarkar\thanks{Co-Author}
  ,\\\textbf{Venkataramana Runkana}\\
  TCS Research \\
  \texttt{sagar.sakhinana@tcs.com,rajat.sarkar1@tcs.com} \\
  \texttt{venkat.runkana@tcs.com}
}

\begin{document}
\maketitle

\vspace{-8mm}
\begin{abstract} 
\vspace{-3mm}
Anomaly detection is fundamental yet, challenging problem with practical applications in industry. The current approaches neglect the higher-order dependencies within the networks of interconnected sensors in the high-dimensional time series(multisensor data) for anomaly detection. To this end, we present a self-adapting anomaly detection framework for joint learning of (a) discrete hypergraph structure and (b) modeling the temporal trends and spatial relations among the interdependent sensors using the hierarchical encoder-decoder architecture to overcome the challenges. The hypergraph representation learning-based framework exploits the relational inductive biases in the hypergraph-structured data to learn the pointwise single-step-ahead forecasts through the self-supervised autoregressive task and predicts the anomalies based on the  forecast error. Furthermore, our framework incentivizes learning the anomaly-diagnosis ontology through a differentiable approach. It derives the anomaly information propagation-based computational hypergraphs for root cause analysis and provides recommendations through an offline, optimal predictive control policy to remedy an anomaly. We conduct extensive experiments to evaluate the proposed method on the benchmark datasets for fair and rigorous comparison with the popular baselines. The proposed method outperforms the baseline models and achieves SOTA performance. We report the ablation studies to support the efficacy of the framework.
\end{abstract}

\vspace{-7mm}
\section{Introduction}
\vspace{-3mm}
Anomaly detection is a long-standing task that has a wide range of applications in monitoring the behaviors of various dynamical real-world systems, including but not limited to financial markets, retail, and e-commerce. Of particular interest in this work is anomaly detection on industrial data for developing a framework that automates anomaly detection, identifies the root causes, and provides recommendations to resolve the underlying issues with ultra-low inference times. At a higher hierarchical level, large-scale industrial units are complex interaction systems of nonlinear dynamical subunits that operate within the margins of normal optimal conditions(NOCs), i.e., anomaly-free. The subunits with a copious amount of interdependent sensors generate high-dimensional time-ordered data capturing the collective behavior of subunits, mostly under NOCs. Anomalies are rare events. Detecting and identifying the anomalies in the large-scale system monitoring data is challenging. Under this scenario, the defacto approach is an unsupervised anomaly detection task on multidimensional data. The goal is to learn the temporal dependencies and the spatial correlations among the interdependent sensors for modeling normality.  Any deviation from such learned relationships suggests the occurrence of abnormality, which is inherently different from the normality patterns. Of late, deep learning-based time series modeling techniques have attracted interest in learning the signatures of abnormality events in high-dimensional time series data. In this vein, traditional deep learning-enabled anomaly detection techniques are the reconstruction-based techniques\cite{li2021multivariate, zhou2019beatgan, su2019robust, zong2018deep, li2019mad}; the forecasting-based techniques\cite{angiulli2002fast, lazarevic2005feature, hundman2018detecting, tariq2019detecting}; the density-estimation techniques\cite{zong2018deep, yairi2017data, breunig2000lof}; the clustering-based techniques\cite{shin2020itad, shen2020timeseries, ruff2018deep}. Despite the success, they neglect the inter-relations among the sensors in modeling the intricate
spatiotemporal dynamics for better root-cause analysis of the anomalies. In the recent past, graph neural networks(GNNs) have emerged successful in representing and analyzing complex graph-structured data and thus have likewise been demonstrated for the multivariate time-series anomaly detection task\cite{zhao2020multivariate, deng2021graph, chen2021learning, zhanggrelen}. Nonetheless, GNNs suffer from inherent limitations of pairwise associations, whereas the relations within the networks of interconnected sensors could go beyond pairwise connections. Hypergraphs offer a natural fit for modeling the higher-order relations underlying the interconnected sensor networks in the multisensor data. The hypergraph structured data has the inherent relational information embedded across a wide-range of hypergraph structural-property characteristics spanning the subhypergraphs. However, the existing Hypergraph Convolutional Neural Networks(HgCNNs\cite{yadati2019hypergcn,feng2019hypergraph}) have inherent drawbacks in their ability to learn the hypergraph structural characteristics for jointly modeling intra-series temporal dependencies and inter-series correlations of the multisensor data. To overcome these challenges, we present the Hypergraph anomaly detection framework(HgAD). Our framework is tailored to model the inherent inter and intra-dependency relations simultaneously among multiple time series of the interconnected-sensor networks for the pointwise predictions on the self-supervised autoregressive task. We detect anomalies based on the pointwise forecast error. In addition, we provide additional insights into the occurrence of abnormal events through a root cause analysis and provide recommendations to offset the anomaly for improved throughput and efficiency $-$ all to drive profitability. Our contributions are summarized as follows. (1) a hypergraph structure learning technique to infer the dynamic hypergraph through the similarity learning-based method for structured spatio-temporal representation of the multisensor data. (2) novel attention-based hypergraph convolution technique to perform spatial-hypergraph filtering operation, (3) novel local-hypergraph pooling and unpooling techniques\cite{gao2019graph} to perform downsampling and upsampling operations on hypergraphs of arbitrary structure, and (4) novel single-stage hierarchical attention-based encoder-decoder technique to learn higher-order representations of the hypergraph structured data. The rest of this paper had organized as follows. Section \ref{sec:approach} presents the working mechanism of our framework. Section \ref{sec:experiments} describes the experimental setup, shows the performance of the proposed method in comparison with the baselines, and discusses the ablation studies. We conclude this work in Section  \ref{sec:conclusion}.

\vspace{-5mm}
\section{Our approach}\label{sec:approach} 
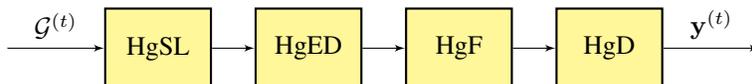
\begin{figure}[htbp]
\center
\begin{tikzpicture}[auto, node distance=2cm,>=latex']
    \node [input, name=input] {};
    \node [sum, right of=input] (sum) {};
    \node [block, thick, right of=sum] (controller) {$\text{HgSL}$};
    \node [block, thick, right of=controller, node distance=2cm] (system) {$\text{HgED}$};
    \node [block, thick, right of=system, node distance=2cm] (system_) {$\text{HgF}$};
    \node [block, thick, right of=system_, node distance=2cm] (system__) {$\text{HgD}$};
    
    \draw [->] (sum) -- node {$\mathcal{G}^{(t)}$} (controller);
    \node [output, right of=system__] (output) {};
     
    \draw [->] (controller) -- node[label={[xshift=0.0cm, yshift=-0.2cm]$ $}] {} (system);
    \draw [->] (system) -- node[label={[xshift=0.0cm, yshift=-0.2cm]$ $}] {} (system_);
    \draw [->] (system_) -- node[label={[xshift=0cm, yshift=-0.2cm]$ $}] {} (system__);
    \draw [->] (system__) -- node[label={[xshift=0cm, yshift=-0.2cm]$ $}] {$\mathbf{y}^{(t)}$} (output);
    
\end{tikzpicture}
\vspace{0mm}
\caption{The $\text{HgAD}$ framework operates on the time-varying hypergraph-structured data $\mathcal{G}^{(t)}$ at time step t and predicts it as normal or not-normal.} \label{fig:a}
\end{figure}

\vspace{-4mm}
Our framework consists of the following modules. (a) The hypergraph structure learning(HgSL) module learns the underlying higher-order structural patterns of the multisensor data that captures the inherent dependency relationships within the network of interconnected sensors. (b) The encoder-decoder(HgED) module operates on the hypergraph topology to learn the hierarchical representations that implicitly encode both the temporal trends and spatial relations among the multiple IoT sensors while preserving the global topological properties of the hypergraph. The module learns the discriminative hypernode-level representations in the (low-dimensional) Euclidean space to distinguish the normal-abnormal instances. (c) The hypergraph forecasting(HgF) module utilizes the hypernode representations and predicts the one-step-ahead forecasts of the sensors. (d) The hypergraph deviation(HgD) module flags the probable anomalous events by comparing the expected and observed trends of the sensors. This module utilizes the forecasting error as the criterion for anomaly detection and serves as supervisory information for the HgSL module. Figure \ref{fig:a} shows the HgAD framework. Furthermore, the framework provides the root cause analysis by viewing the learned hypergraph topology as a computation hypergraph for anomaly information traversal across the interconnected network of sensors.  The knowledge of the inherent structure plausibly explains the manifestation of the underlying cause when the predicted relationships deviate from the learned normality relationships in the hypergraph-structured data. The objective of unsupervised anomaly detection is to predict the output labels $\mathbf{y}^{(t)} \in\{0,1\}$, which is a two-label prediction task suggesting the anomaly occurrence at a time point, $t$. Note : $\text{label}=\text{1}$ means \enquote{anomaly} and $\text{label}=\text{0}$ means \enquote{normal}.

\vspace{-4mm}

\vspace{-5mm}
\subsection{Hypergraph Representation}
\begin{figure}[htbp]
\center
\begin{tikzpicture}[scale=0.55]
        \node (v1) at (0, 2) {};
        \node (v2) at (1.5, 3) {};
        \node (v3) at (4, 2.5) {};
        \node (v4) at (0, 0) {};
        \node (v5) at (2, 0.5) {};
        \node (v6) at (3.5, 0) {};

        \begin{scope}[fill opacity = 0.8]
            \filldraw [fill = yellow!70] ($(v1) + (-0.5, 0)$)
            to [out = 90, in = 180] ($(v2) + (0, 0.5)$)
            to [out = 0,in = 90] ($(v3) + (1, 0)$)
            to [out = 270, in = 0] ($(v2) + (1, -0.8)$)
            to [out = 180, in = 270] ($(v1) + (-0.5, 0)$);
            \filldraw [fill = blue!50] ($(v4) + (-0.5, 0.2)$)
            to [out = 90, in = 180] ($(v1) + (0, 1)$)
            to [out = 0, in = 90] ($(v4) + (0.6, 0.3)$)
            to [out = 270, in = 0] ($(v4) + (0, -0.6)$)
            to [out = 180, in = 270] ($(v4) + (-0.5, 0.2)$);
            \filldraw [fill = green!70] ($(v5) + (-0.5, 0)$)
            to [out = 90, in = 225] ($(v3) + (-0.5, -1)$)
            to [out = 45, in = 270] ($(v3) + (-0.7, 0)$)
            to [out = 90, in = 180] ($(v3) + (0, 0.5)$)
            to [out = 0, in = 90] ($(v3) + (0.7, 0)$)
            to [out = 270, in = 90] ($(v3) + (-0.3, -1.8)$)
            to [out = 270, in = 90] ($(v6) + (0.5, -0.3)$)
            to [out = 270, in = 270] ($(v5) + (-0.5, 0)$);
            \filldraw [fill = red!70] ($(v2) + (-0.5, -0.2)$)
            to [out = 90, in = 180] ($(v2) + (0.2, 0.4)$)
            to [out = 0, in = 180] ($(v3) + (0, 0.3)$)
            to [out = 0, in = 90] ($(v3) + (0.3, -0.1)$)
            to [out = 270, in = 0] ($(v3) + (0, -0.3)$)
            to [out = 180, in = 0] ($(v3) + (-1.3, 0)$)
            to [out = 180, in = 270] ($(v2) + (-0.5, -0.2)$);
            \filldraw [fill = purple!50] ($(v5) + (-0.5, -0.2)$)
            to [out = 90, in = 180] ($(v5) + (0.2, 0.4)$)
            to [out = 0, in = 180] ($(v6) + (0, 0.3)$)
            to [out = 0, in = 90] ($(v6) + (0.3, -0.1)$)
            to [out = 270, in = 0] ($(v6) + (0, -0.3)$)
            to [out = 180, in = 0] ($(v6) + (-1.3, 0)$)
            to [out = 180, in = 270] ($(v5) + (-0.5, -0.2)$);         
        \end{scope}
        \begin{pgfonlayer}{background}
        \draw[edge,color=orange] (v4) -- (v5);
        \end{pgfonlayer}

        \foreach \i in {1, 2, ..., 6}
        {
            \fill (v\i) circle (0.1);
        }

        \fill (v1) circle (0.1) node [right] {$v_1$};
        \fill (v2) circle (0.1) node [below right] {$v_2$};
        \fill (v3) circle (0.1) node [left] {$v_3$};
        \fill (v4) circle (0.1) node [below] {$v_4$};
        \fill (v5) circle (0.1) node [below right] {$v_5$};
        \fill (v6) circle (0.1) node [below left] {$v_6$};
        
        \begin{scope}[every node/.style = {fill, shape = circle, node distance = 12pt}]
            \node (e1) [color = yellow!56, label = right:$e_1$] at (-3, 3) {};
            \node (e2) [below of = e1, color = red!56, label = right:$e_2$] {};
            \node (e3) [below of = e2, color = green!56, label = right:$e_3$] {};
            \node (e4) [below of = e3, color = blue!56, label = right:$e_4$] {};
            \node (e5) [below of = e4, color = orange, label = right:$e_5$] {};
            \node (e6) [below of = e5, color = purple!56, label = right:$e_6$] {};
        \end{scope}
                
\end{tikzpicture}
\vspace{-2mm}
\caption{The hypernodes represent the multiple IoT sensors(denoted by $v_{i}$). The hyperedges(denoted by $e_{i}$) connect an arbitrary number of hypernodes.}
\label{fig:b}
\end{figure}
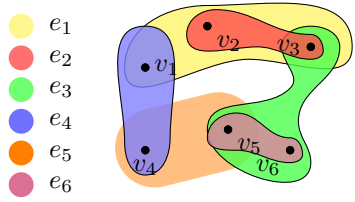

\vspace{-2mm}
Consider a multivariate contiguous time-series data from $n$ variates(sensors) observed over $\mathrm{T}$ time steps denoted  by $\mathbf{f} = \big[\mathbf{f}^{(1)}, \ldots, \mathbf{f}^\mathrm{(T)}  \big]$. $\mathbf{f}^{(t)} \in \mathbb{R}^{n}$ represents the $n$-variates observations at time step t, which form an $n$-dimensional vector. We generate fixed-length inputs by a sliding  window of length $w$ over historical data as input to our framework to learn from the multisensor data. The windowed data at time $t$, $\mathbf{F}^{(t)} \in \mathbb{R}^{n \times w}$ is given by $\mathbf{F}^{(t)} = \big[\mathbf{f}^{(t-w)}, \mathbf{f}^{(t-w+1)}, \ldots, \mathbf{f}^{(t-1)} \big]$. The interconnected sensors do not have an explicit higher-order relational structure underlying the multisensor data. We utilize a hypergraph framework for a structured representation of the data. The hypergraph structured data offers a natural way to abstract the complex interdependencies among the sensors. We denote the sensors as the hypernodes of the dynamic bidirected hypergraph. The bidirectional hyperedges model the super-dyadic relations among the hypernodes. We obtain spatiotemporal hypergraph data with time-varying hypernode features. The hypernode, hyperedge set, and hypergraph structure remain unchanged.  Figure \ref{fig:b} illustrates the hypergraph representation of the network of interconnected sensors.

\vspace{-6mm}

\vspace{-4mm}
We observe a hypergraph at the time-step $t$ ($w+1 \leq t \leq T$) and is defined as $\mathcal{G}^{(t)}=(\mathcal{V}, \mathcal{E}, \mathbf{F}^{(t)})$ where the hypernode set of $\mathcal{V}=\left\{v_{1}, v_{2}, \ldots, v_{n}\right\}$ is the set of $n$ hypernodes, and the hyperedge set of $\mathcal{E}=\left\{e_{1}, e_{2}, \ldots, e_{m}\right\}$ is a set of $m$ hyperedges, and the $\mathbf{F}^{(t)} \in \mathbb{R}^{n \times w}$ is the hypernode-feature matrix. The incidence matrix, $\mathbf{H} \in \mathbb{R}^{n \times m}$, describes the relational structure in the hypergraph. $\mathbf{H}_{i, \hspace{0.5mm}p}=1$ if the hyperedge $p$ contains the hypernode $i$ and otherwise 0. The incidence matrix $\mathbf{H}$ is symmetric for the bidirected hypergraph. Let $\mathcal{N}_{p}=\big\{v_{i} | \mathbf{H}_{i, p} = 1\big\}$ represents the subset of hypernodes $v_{i} \in \mathcal{V}$ belonging to hyperedge $p$. The local neighborhood of the incident hypernode $i$ is given by $\mathcal{N}_{p} \backslash i$. The global neighborhood of hypernode $i$, $\mathcal{N}_{i} = \big\{e_{p}|\mathbf{H}_{i, p} = 1\big\}$, which contains the set of hyperedges $e_{p}$ incident with hypernode $i$. Given a historical time series data over $w$ time steps of $n$-variates $\mathbf{F}^{(t)} \in \mathbb{R}^{n \times w}$, the goal is to predict the occurrence of an anomaly at $t$ and explain the cause of deviations.

\vspace{-4mm}
\subsection{Hypergraph Structure Learning(HgSL)}\label{hgsl} 
\vspace{-2mm}
The traditional methods\cite{feng2019hypergraph, yadati2019hypergcn} operate on the prior known explicit structure  to learn from the hypergraph-structured data. In practice, however, the real-world applications present challenging scenarios where the higher-order structural dependencies in networks underlying the interdependent sensors is unknown, incomplete, or partially available due to limited prior knowledge. The HgSL module offers a structural modeling approach to dynamically optimize the sensor topology and implicitly learn the task-relevant relational structure from the hypernode(sensor) embeddings. The optimal sensor topology captures the complex hidden relations among the interdependent sensors. We perform inference on the hypergraph-structured multisensor data through the hypergraph representation learning on the downstream anomaly detection task driven by the inductive-learning approach. The hypernodes of the hypergraph had characterized by learnable (low-dimensional) embeddings $\mathbf{z}_{i}$, $1 \leq i \leq n$. The hypernode embeddings $\mathbf{z}_{i} \in \mathbb{R}^{d}$ are continuous vector representations in the $d$-dimensional embedding space. We will form a directed hyperedge from  hypernode $i$ by connecting with its $k$-nearest hypernodes. In essence, the hyperedge connects the hypernodes $i$ and $j$; if $j$ is among the $k$-closest local-hypergraph neighbors of $i$. We obtain $n$ (in this setting, m = n) hyperedges that are incident with $k+1$ non-repeating hypernodes of the hypergraph. We accomplish this by learning a sparse symmetric hypergraph incidence matrix $\mathbf{H}$ in which $n \times (k+1)$ elements are equal to 1 while the rest are zero. We achieve this by computing the pairwise Euclidean distance of the hypernodes using their embeddings to learn the similarity measure. It is described by,

\vspace{-5mm}
\begin{equation}
d_{i, j} = \big(\sum_{ind=1}^{d}{|\mathbf{z}^{ind}_{i} - \mathbf{z}^{ind}_{j}|^{2}}\big)^{1/2}; \hspace{1mm} i \in \{1, \ldots, n\},  i \neq j
\label{eq:hgsl1}
\end{equation}

\vspace{-2mm}
where $d_{i, j}$ denotes the Euclidean distance between a pair of hypernodes, $i$, and $j$ in embedding space. We obtain a static bidirected hypergraph structure(nearest-neighbor hypergraph), in which each hyperedge encodes the relationship between the hypernode and its $k$-nearest local-neighbor hypernodes by connecting them according to the incidence matrix $\mathbf{H} \in \mathbb{R}^{n \times n}$ described below,

\vspace{-5mm}
\begin{equation}
\mathbf{H}_{j,p} = \mathbbm{1}\big\{j \in \operatorname{Topk}\big(\operatorname{min}\big\{d^{(p)}_{i,j}\big\}, j \in \mathcal{V}\big)\big\} | \mathbf{H}_{i,p}=1
\label{eq:hgsl2}
\end{equation}

\begin{figure}[htbp]
    \hspace{-15mm}\includegraphics[width=1.25\textwidth]{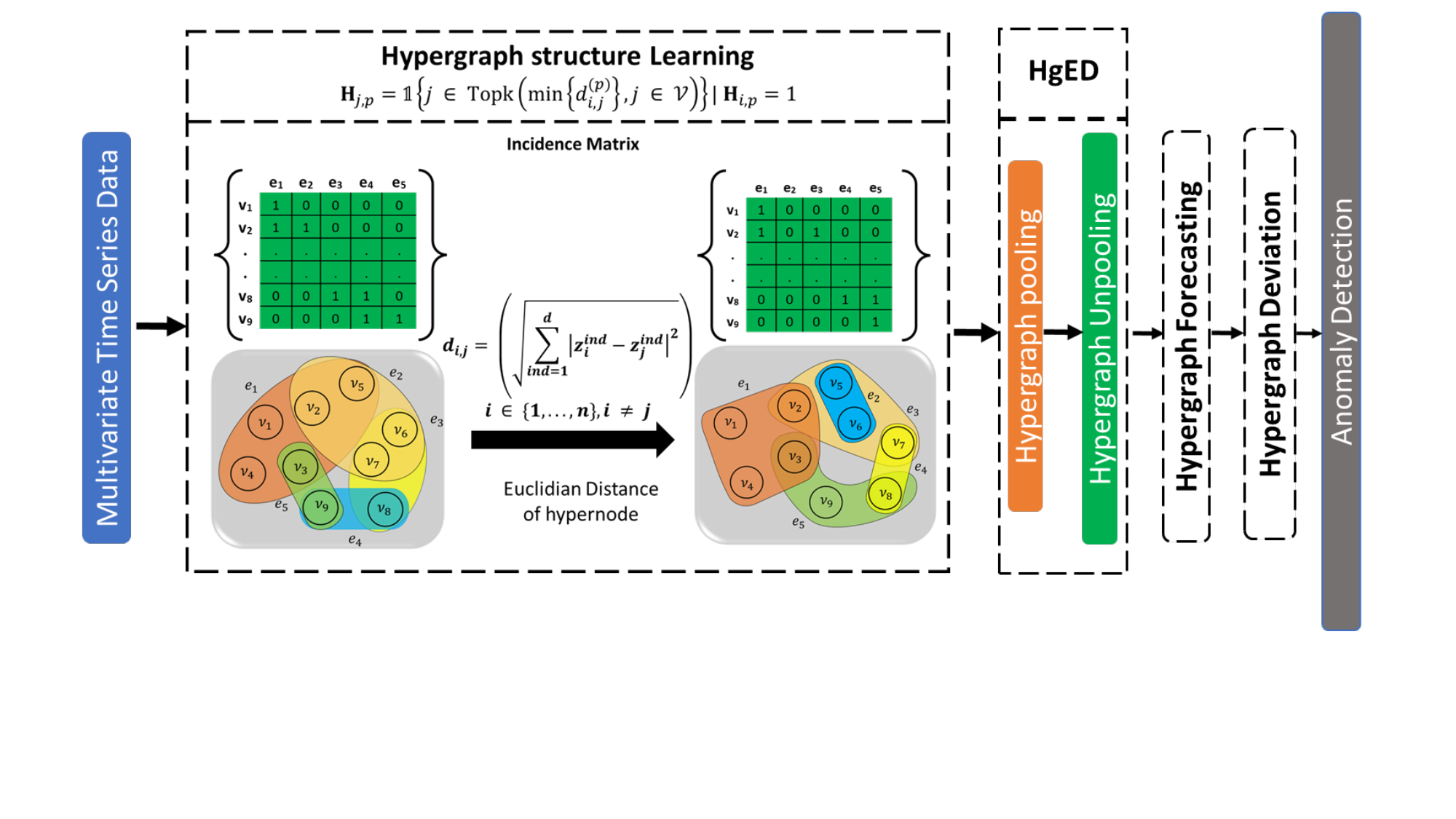}
    \vspace{-30mm}
    \caption{Overview of the HgAD framework. The differentiable Hypergraph Structure Learning(HgSL) module learns the discrete optimal hypergraph structure to facilitate the downstream anomaly detection. The HgED module learns the higher-order dependencies and hypergraph forecasting module further distills the knowledge for lower forecasting error. The Hypergraph-Deviation(HgD) module flags the anomalous behaviour based on single time-point forecast error.}
    \label{fig:a(HgSL)}
\end{figure}

\vspace{-1mm}
where the mathematical operator $\operatorname{Topk}$ returns the indices of the $k$-nearest hypernodes of $i$. The hyperparameter $k$ controls the sparsity of the hypergraph. The HgSL module learns the hidden $k$-uniform hypergraph structure underlying the sensors for modeling the dynamic sensory data. Figure \ref{fig:a(HgSL)} illustrates the HgSL module along with the downstream modules.

\vspace{-3mm}

\vspace{-4mm} 
\subsection{Hypernode Positional Encoding(HPE)}\label{hpe} 
\vspace{-2mm}
There exists no canonical ordering of the hypernodes in the hypergraph. We encode hypernode positional information to enable position awareness\cite{chen2021learning}. The positional embeddings, $\text{P}_{E} \in \mathbb{R}^{n \times w}$, have been linearly added to the feature matrix $\mathbf{F}^{(t)} \in \mathbb{R}^{n \times w}$ to inform the model about the positional information of the hypernodes from the main hypergraph. The HPE module generates the positional embeddings based on the local and global neighbors of each hypernode.

\vspace{-3mm}
\subsection{Hypergraph Encoder-Decoder(HgED)}\label{HgEnc-Dec}
\vspace{-2mm}
We will first discuss the Hypergraph Convolutional Neural Network(HgCNN), Hypergraph Local-Pooling(HgPool), and Hypergraph UnPooling(UnHgPool) operators. We then elaborate on the HgED module in great detail.

\vspace{-1mm}
\subsubsection{Hypergraph Convolutional Neural Network(HgCNN)}\label{hgcnn} 
\vspace{-2mm}
The HgCNN operator generalizes the convolution operation to the hypergraphs. It presents the neural network primitives that compute the hypernode representations $x_{v_{i}}$, $1 \leq i \leq n$, where $x_{v_{i}} \in \mathbb{R}^{d}$.  We obtain hypernode-level representations through propagating, aggregating, and transforming the hypernode-level feature information on the hypergraph topology to learn from the sensor-based information. It performs a spatial-hypergraph filtering operation in two phases. At first, given the incidence matrix $\mathbf{H} \in \mathbb{R}^{n \times n}$  and hypernode feature matrix $\mathbf{F}^{(t)} \in \mathbb{R}^{n \times w}$. We determine the hyperedge attribute matrix $\mathbf{X}^{(t)}_{e} \in \mathbb{R}^{n \times d}$. The row-vectors in $\mathbf{X}^{(t)}_{e}$ denote the hyperedge representations $x^{(t)}_{e_{p}} \in \mathbb{R}^{d}, 1 \leq p \leq n$ and are computed by aggregating the attention-weighted features of the incident hypernodes obtained as,

\vspace{-4mm}
\begin{equation}
\normalsize{
\begin{aligned}
x^{(t)}_{e_{p}} = \sigma \big(\hspace{-1mm}\sum_{i \hspace{0.5mm}\in \hspace{0.5mm}{\mathcal{N}_{p}}} \hspace{-1mm} \alpha_{p, i} \mathbf{W}_{1}\mathbf{F}^{(t)}_{i}\big) 
\end{aligned}
} \label{eq:hgcnn1}
\end{equation}

\vspace{-4mm}
where $\mathbf{F}^{(t)}_{i} \in \mathbb{R}^{w}$ denotes the feature vector of hypernode $i$ and $\mathbf{W}_{1} \in \mathbb{R}^{d \times w}$ is the learnable weight matrix. $\sigma$ is the sigmoid function to introduce non-linearity in the intra-neighborhood feature aggregation scheme. The hyperedge representations embeds the latent information of the intra-relations among the incident hypernodes. We utilize the hypernode embedding vectors($\mathbf{z}_{i}$) to learn the attention score $e_{p, i}$ of hyperedge $p$ incident with the hypernode $i$. The normalized attention score $\alpha_{p, i}$ determined using the softmax function is computed by,

\vspace{-6mm}
\begin{gather}
e_{p, i} =\operatorname{ReLU}\big(\mathbf{W}^{\top}_{2} \cdot \mathbf{g}_{i}^{(t)}\big) ; \mathbf{g}_{i}^{(t)} =\mathbf{z}_{i} \oplus \mathbf{W}_{1}\mathbf{F}^{(t)}_{i} \label{eq:hgcnn3} \\ 
\alpha_{p, i} =\frac{\exp \left(e_{p, i}\right)}{\sum_{i \hspace{0.5mm}\in \hspace{0.5mm}{\mathcal{N}_{p}}} \exp \left(e_{p, i}\right)} \nonumber
\end{gather}  

\vspace{-3mm}
where $\mathbf{W}_{2} \in \mathbb{R}^{2d}$ is the learnable vector. $\oplus$ denotes the concatenation operation. The next phase captures the complex non-linear inter-relations between the hypernodes and hyperedges. We perform the hypergraph attention-based global neighborhood aggregation scheme for learning informative and expressive hypernode representations. In simple terms, the hyperedge attribute extractor fuses the hyperedge information with the incident hypernodes as described by,

\vspace{-3mm}
\begin{equation}
x^{(t)}_{v_{i}}=\operatorname{ReLU}\big(\mathbf{W}_{3}\mathbf{F}^{(t)}_{i} + \sum_{p \in \mathcal{N}_{i}} \beta_{i, p} \mathbf{W}_{4} x^{(t)}_{e_{p}}\big)
\label{eq:hgcnn2}
\end{equation}

\vspace{-3mm}
where $\mathbf{W}_{3} \in \mathbb{R}^{d \times w}$ and $\mathbf{W}_{4} \in \mathbb{R}^{d \times d}$ are learnable weight matrices. We utilize the $\operatorname{ReLU}$ activation function to introduce non-linearity to update hypernode-level representations. The unnormalized attention score of hypernode $i$ incident with hyperedge $p$  given by $\phi(i, p)$, and the attention coefficients $\beta_{i, p}$ computed as,

\vspace{-8mm}
\begin{gather}
\phi(i, p) = \operatorname{ReLU}\big(\mathbf{W}_{5}^{\top} \cdot \big(\mathbf{g}_{i}^{(t)} \oplus \mathbf{W}_{4} x^{(t)}_{e_{p}}\big)\big)  \nonumber \\
\beta_{i, p} = \frac{\exp (\phi(i, p))}{\sum_{p \hspace{0.5mm}\in \hspace{0.5mm}{\mathcal{N}_{i}}} \exp (\phi(i, p))} \nonumber 
\end{gather}

\begin{figure}[htbp]
    \hspace{-20.0mm}\includegraphics[width=1.3\textwidth]{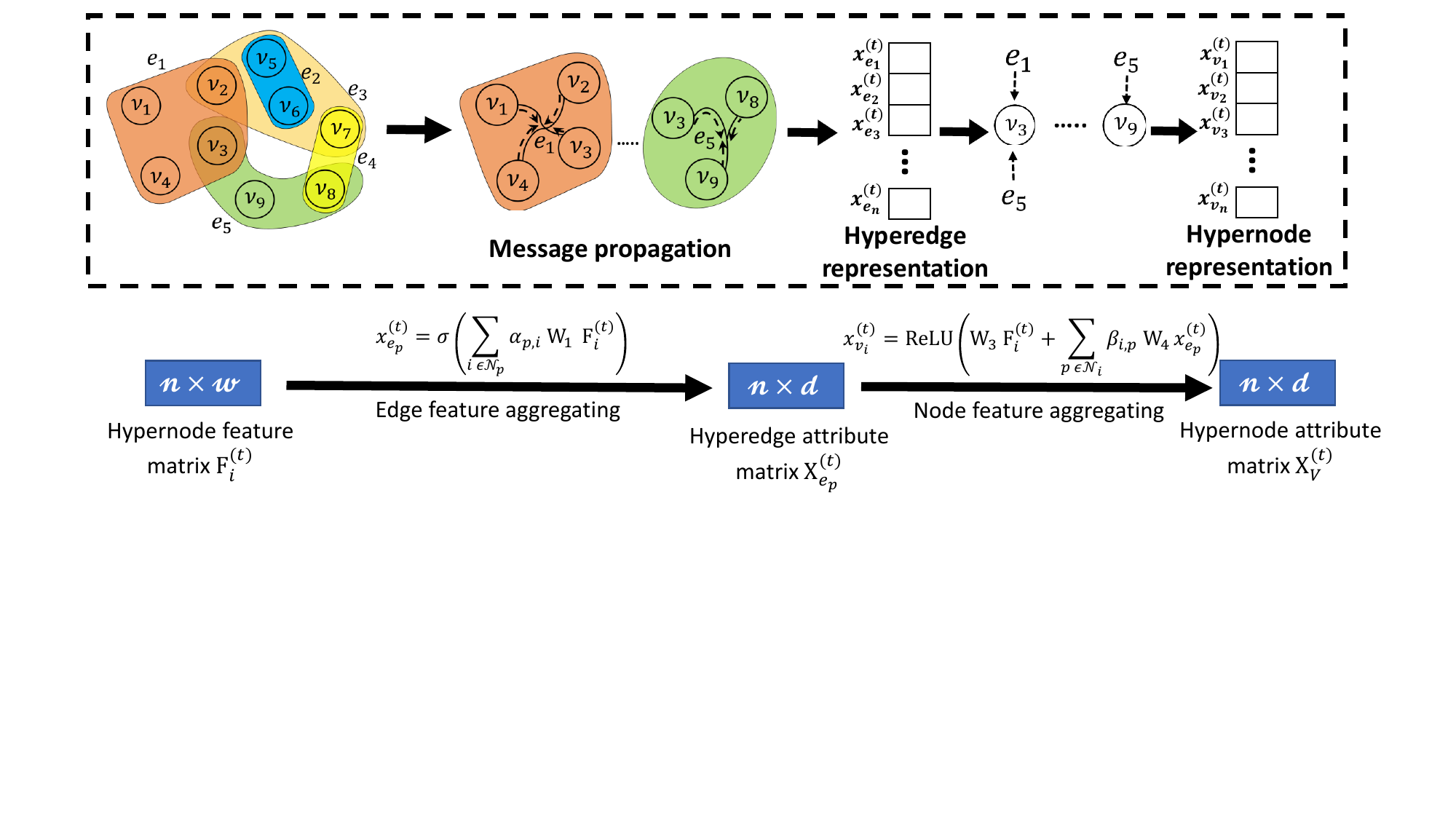}
    \vspace{-45.0mm}
    \caption{The HgCNN operator performs the neural message-passing schemes on the hypergraphs.}
    \label{fig:b(HgCNN)}
\end{figure}

\vspace{-2mm}
where $\mathbf{W}_{5} \in \mathbb{R}^{3d}$ is a trainable vector. In summary, the HgCNN operator performs the neighborhood aggregation schemes to learn the optimal hypernode-level representations, whilst maximally preserving the high-order relations embedded in the structural characteristics of the hypergraph. Figure \ref{fig:b(HgCNN)} illustrates the HgCNN module.
 
\vspace{-4mm}

\vspace{-5mm}
\begin{figure}[htbp]
\centering
\begin{minipage}{.35\linewidth}
\centering
\begin{tikzpicture}[scale=0.6]
\begin{scope}[yshift=-6cm]
        \node (v1) at (0, 2) {};
        \node (v2) at (1.5, 3) {};
        \node (v3) at (4, 2.5) {};
        \node (v4) at (0, 0) {};
        \node (v5) at (2, 0.5) {};
        \node (v6) at (3.5, 0) {};

        \begin{scope}[fill opacity = 0.8]
            \filldraw [fill = yellow!70] ($(v1) + (-0.5, 0)$)
            to [out = 90, in = 180] ($(v2) + (0, 0.5)$)
            to [out = 0,in = 90] ($(v3) + (1, 0)$)
            to [out = 270, in = 0] ($(v2) + (1, -0.8)$)
            to [out = 180, in = 270] ($(v1) + (-0.5, 0)$);
            \filldraw [fill = blue!50] ($(v4) + (-0.5, 0.2)$)
            to [out = 90, in = 180] ($(v1) + (0, 1)$)
            to [out = 0, in = 90] ($(v4) + (0.6, 0.3)$)
            to [out = 270, in = 0] ($(v4) + (0, -0.6)$)
            to [out = 180, in = 270] ($(v4) + (-0.5, 0.2)$);
            \filldraw [fill = green!70] ($(v5) + (-0.5, 0)$)
            to [out = 90, in = 225] ($(v3) + (-0.5, -1)$)
            to [out = 45, in = 270] ($(v3) + (-0.7, 0)$)
            to [out = 90, in = 180] ($(v3) + (0, 0.5)$)
            to [out = 0, in = 90] ($(v3) + (0.7, 0)$)
            to [out = 270, in = 90] ($(v3) + (-0.3, -1.8)$)
            to [out = 270, in = 90] ($(v6) + (0.5, -0.3)$)
            to [out = 270, in = 270] ($(v5) + (-0.5, 0)$);
            \filldraw [fill = red!70] ($(v2) + (-0.5, -0.2)$)
            to [out = 90, in = 180] ($(v2) + (0.2, 0.4)$)
            to [out = 0, in = 180] ($(v3) + (0, 0.3)$)
            to [out = 0, in = 90] ($(v3) + (0.3, -0.1)$)
            to [out = 270, in = 0] ($(v3) + (0, -0.3)$)
            to [out = 180, in = 0] ($(v3) + (-1.3, 0)$)
            to [out = 180, in = 270] ($(v2) + (-0.5, -0.2)$);
            \filldraw [fill = purple!50] ($(v5) + (-0.5, -0.2)$)
            to [out = 90, in = 180] ($(v5) + (0.2, 0.4)$)
            to [out = 0, in = 180] ($(v6) + (0, 0.3)$)
            to [out = 0, in = 90] ($(v6) + (0.3, -0.1)$)
            to [out = 270, in = 0] ($(v6) + (0, -0.3)$)
            to [out = 180, in = 0] ($(v6) + (-1.3, 0)$)
            to [out = 180, in = 270] ($(v5) + (-0.5, -0.2)$);         
        \end{scope}
        \begin{pgfonlayer}{background}
        \draw[edge,color=orange] (v4) -- (v5);
        \end{pgfonlayer}

        \foreach \i in {1, 2, ..., 6}
        {
            \fill (v\i) circle (0.1);
        }

        \fill (v1) circle (0.1) node [right] {$v_1$};
        \fill (v2) circle (0.1) node [below right] {$v_2$};
        \fill (v3) circle (0.1) node [left] {$v_3$};
        \fill (v4) circle (0.1) node [below] {$v_4$};
        \fill (v5) circle (0.1) node [below right] {$v_5$};
        \fill (v6) circle (0.1) node [below left] {$v_6$};
\end{scope}   
\end{tikzpicture}
\end{minipage}%
\hspace{-4cm}\begin{minipage}{.5\linewidth}
\centering
\begin{tikzpicture}
  \coordinate (A) at (0,0);
  \coordinate (B) at (2,0);
  \draw[->] (A) -- (B) node[midway,fill=white] {\text{HgPool}};
\end{tikzpicture}
\end{minipage}%
\hspace{-4cm}\begin{minipage}{.45\linewidth}
\centering
\begin{tikzpicture}[scale=0.6]
\begin{scope}[yshift=-6cm]
        \node (v1) at (0, 2) {};
        \node (v3) at (4, 2.5) {};
        \node (v4) at (0, 0) {};
        \node (v5) at (2, 0.5) {};
        \node (v6) at (3.5, 0) {};

        \begin{scope}[fill opacity = 0.8]
            \filldraw [fill = yellow!70] ($(v1) + (-0.5, 0)$)
            to [out = 90, in = 180] ($(v3) + (0, 0.5)$)
            to [out = 0,in = 90] ($(v3) + (1, 0)$)
            to [out = 270, in = 0] ($(v1) + (1, -0.8)$)
            to [out = 180, in = 270] ($(v1) + (-0.5, 0)$);
            \filldraw [fill = blue!50] ($(v4) + (-0.5, 0.2)$)
            to [out = 90, in = 180] ($(v1) + (0, 1)$)
            to [out = 0, in = 90] ($(v4) + (0.6, 0.3)$)
            to [out = 270, in = 0] ($(v4) + (0, -0.6)$)
            to [out = 180, in = 270] ($(v4) + (-0.5, 0.2)$);
            \filldraw [fill = green!70] ($(v5) + (-0.5, 0)$)
            to [out = 90, in = 225] ($(v3) + (-0.5, -1)$)
            to [out = 45, in = 270] ($(v3) + (-0.7, 0)$)
            to [out = 90, in = 180] ($(v3) + (0, 0.5)$)
            to [out = 0, in = 90] ($(v3) + (0.7, 0)$)
            to [out = 270, in = 90] ($(v3) + (-0.3, -1.8)$)
            to [out = 270, in = 90] ($(v6) + (0.5, -0.3)$)
            to [out = 270, in = 270] ($(v5) + (-0.5, 0)$);
  
        \end{scope}
        \begin{pgfonlayer}{background}
        \draw[edge,color=orange] (v4) -- (v5);
        \end{pgfonlayer}

        \foreach \i in {1, 3, ..., 5}
        {
            \fill (v\i) circle (0.1);
        }

        \fill (v1) circle (0.1) node [right] {$v_1$};
        \fill (v3) circle (0.1) node [left] {$v_3$};
        \fill (v4) circle (0.1) node [below] {$v_4$};
        \fill (v5) circle (0.1) node [below right] {$v_5$};
\end{scope}               
\end{tikzpicture}
\end{minipage}\par\medskip
\vspace{-3mm}
\caption{The HgPool operator performs a downsampling operation. For illustration purposes, we rejected the hypernodes($v_2, v_6$) and the corresponding hyperedges($e_2, e_6$) in the hypergraph(left) to obtain the pooled hypergraph(right).}
\label{fig:c}
\end{figure}
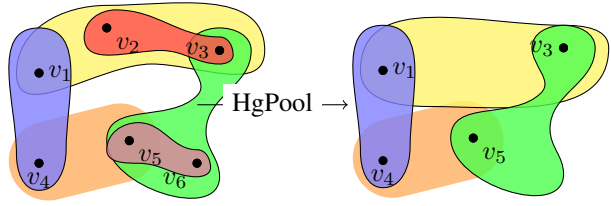 
 
\vspace{0mm}
\subsubsection{Hypergraph Local-Pooling(HgPool)}\label{hgpool}
\vspace{-2mm}
We present a two-fold approach-based differentiable local-hypergraph pooling operator to (1) induce subhypergraphs by reducing the order and size of the main hypergraph to obtain a pooled hypergraph and (2) learn the hierarchical representations of the hypernodes by encoding the dominant structural characteristics of the hypergraph. The prominent steps in a downsampling technique based local-hypergraph pooling mechanism are (a) computing the measure for performing down-sampling on the hypergraph to obtain the pooled hypergraph by dropping fewer hypernodes of lower importance and the corresponding hyperedges, (b) learning the pooled hypergraph connectivity, and then (c) determining the hypernode-level representations of the pooled hypergraph. The HgPool operator assigns a score to all the hypernodes and selects a subset of high-scoring hypernodes to obtain a pooled hypergraph. We learn the high-order representations by performing the neighborhood aggregation schemes on the topology of the pooled hypergraph. Figure \ref{fig:c} illustrates the local-hypergraph pooling operations. Initially, we perform the convolution operation on the hypergraph. The objectives are bifold. Firstly, to capture local regularities. Next, to encode the relations between the hypernodes and hyperedges in the hypernode-level representations. We compute the real-valued hypernode scores as described by,

\vspace{-6mm}
\begin{align}
\mathbf{X}^{(t)}_{v} &= \textrm{HgCNN}\big(\mathbf{F}^{(t)}, \mathbf{H}\big) \label{eq:hgpool1} \\
\mathbf{s}^{(t)} &=  \mathrm{softmax}\big(\theta_{p} * \mathbf{X}^{(t)}_{v} \big) \label{eq:hgpool2}
\end{align}

\vspace{-3mm}
where $\mathbf{s}^{(t)}=[\mathbf{s}^{(t)}_1, \mathbf{s}^{(t)}_2, \ldots, \mathbf{s}^{(t)}_{|\mathcal{V}|}]$. The hypernode score $\mathbf{s}^{(t)}_i$ measure the importance of hypernode $i$ at time step, $t$. $\theta_{p} \in \mathbb{R}^{d}$ applies a shared linear transformation of the hypernode-level representations to compute the importance score of the hypernodes. Let $*$ denote the matrix multiplication. The $\mathrm{softmax}$ operator is the normalized exponential function. The high-ranking hypernodes have complex temporal dependencies and strong-causal relations with neighboring hypernodes. In contrast, the hypernodes with low scores have weak temporal and spatial-dependency patterns. Henceforth, the high-scoring hypernodes capture the prominent hierarchical structural information compared to low-scoring hypernodes. For a given pooling ratio, $p_r \in (0,1]$ and the size of the hypergraph $|\mathcal{V}|$, we determine the number of hypernodes($n_p$) to retain in the pooled hypergraph as given by $n_p = p_r|\mathcal{V}|$. We then select a subset of top \hspace{0.5mm}$n_p$-ranked hypernodes through the hypernode ranking operation by utilizing the scalar scores $\mathbf{s}^{(t)}$ as described by $\mbox{idx}^{(t)} = \mbox{rank}(\mathbf{s}^{(t)}, n_p)$. $\mbox{idx}^{(t)}$ describes the indices of the top \hspace{0.5mm}$n_p$-scoring hypernodes in the hypergraph. This operation rejects $|\mathcal{V}|$ $-p_r|\mathcal{V}|$ hypernodes and their corresponding hyperedges. In essence, selecting the indices in the hypergraph through the hypernode-ranking operation retains the dominant hypernodes from the main hypergraph $\mathcal{G}^{(t)}$ in the pooled hypergraph $\mathcal{\bar{G}}^{(t)}$. The incidence matrix of the pooled hypergraph $\mathbf{\bar{H}}^{(t)}_{p} = \mathbf{H}^{(t)}(\mbox{idx}^{(t)}, \mbox{idx}^{(t)})$, where $\mathbf{\bar{H}}^{(t)}_{p} \in \mathbb{R}^{n_p\times n_p}$ and $\mathbf{H}^{(t)} \in \mathbb{R}^{n\times n}$. The hypernode-level representations of the pooled hypergraph $\mathcal{\bar{G}}^{(t)}$ are obtained as $\tilde{\mathbf{X}}^{(t)}_{v} = \mathbf{X}^{(t)}_{v}(\mbox{idx}, :)$ where $\tilde{\mathbf{X}}^{(t)}_{v} \in \mathbb{R}^{n_p\times d}$ and $\mathbf{X}^{(t)}_{v} \in \mathbb{R}^{n\times d}$. We perform a gating operation to regulate the information flow from the main hypergraph to the pooled hypergraph to avoid over-smoothing of the hypernode representations as described by, 

\vspace{-6mm}
\begin{align}
\tilde{\mathbf{s}}^{(t)} &= \mathrm{ReLU} \big(\mathbf{s}^{(t)}(\mbox{idx})\big) \label{eq:hgpool3}  \\
\mathbf{\bar{X}}^{(t)}_{v} &= \tilde{\mathbf{X}}^{(t)}_{v} \odot \big(\tilde{\mathbf{s}}^{(t)} \mathbf{1}_{d}^{T}\big) \label{eq:hgpool4}
\end{align}

\vspace{-2mm}
where the element-wise multiplication is denoted by $\odot$. $\mathbf 1_{d} \in \mathbb{R}^{d}$ is a vector of size $d$ with each component value of 1. We obtain the hyperedge-level representations of the scaled-down hypergraph by sum-pooling the incident hypernodes representations as given by,

\vspace{-6mm}
\begin{gather}
\bar{x}^{(t)}_{e_{p}} = \sigma \big( \hspace{-1mm}\sum_{i \hspace{0.5mm}\in \hspace{0.5mm}{\bar{\mathcal{N}}_{p}}} \hspace{-1mm} \mathbf{W}_{6}\bar{x}^{(t)}_{v_{i}}\big)  \nonumber 
\end{gather}
 
\vspace{-2mm} 
We utilize global neighborhood aggregation schemes on the pooled hypergraph to capture the inherent high-order relations in the hypernode-level representations. The hierarchical hypernode representations are determined by aggregating the incident hyperedges information as described below,

\vspace{-8mm}
\begin{gather}
\bar{\mathbf{g}}_{i}^{(t)} = \mathbf{z}_{i} \oplus \mathbf{W}_{6} \bar{x}^{(t)}_{v_{i}}; \bar{\mathbf{g}}_{p}^{(t)} = \mathbf{W}_{7} \bar{x}^{(t)}_{e_{p}} \label{eq:hgpool7} \\
\kappa_{i, p} = \operatorname{ReLU}\big(\mathbf{W}_{8}^{\top} \cdot \big(\bar{\mathbf{g}}_{i}^{(t)} \oplus \bar{\mathbf{g}}_{p}^{(t)}\big)\big) \nonumber \\
\zeta_{i, p} = \frac{\exp (\kappa_{i, p})}{\sum_{p \hspace{0.5mm}\in \hspace{0.5mm}{\mathcal{N}_{i}}} \exp (\kappa_{i, p})} \nonumber \\ 
\bar{x}^{(t)}_{v_{i}} = \sigma \big(\mathbf{W}_{6} \bar{x}^{(t)}_{v_{i}} + \sum_{p \hspace{0.5mm}\in {\mathcal{N}_{i}}}  \zeta_{i, p} \mathbf{W}_{7} \bar{x}^{(t)}_{e_{p}}\big)  \label{eq:hgpool5} 
\end{gather}

\begin{figure}[htbp]
    \hspace{-2mm}\includegraphics[width=1.05\textwidth]{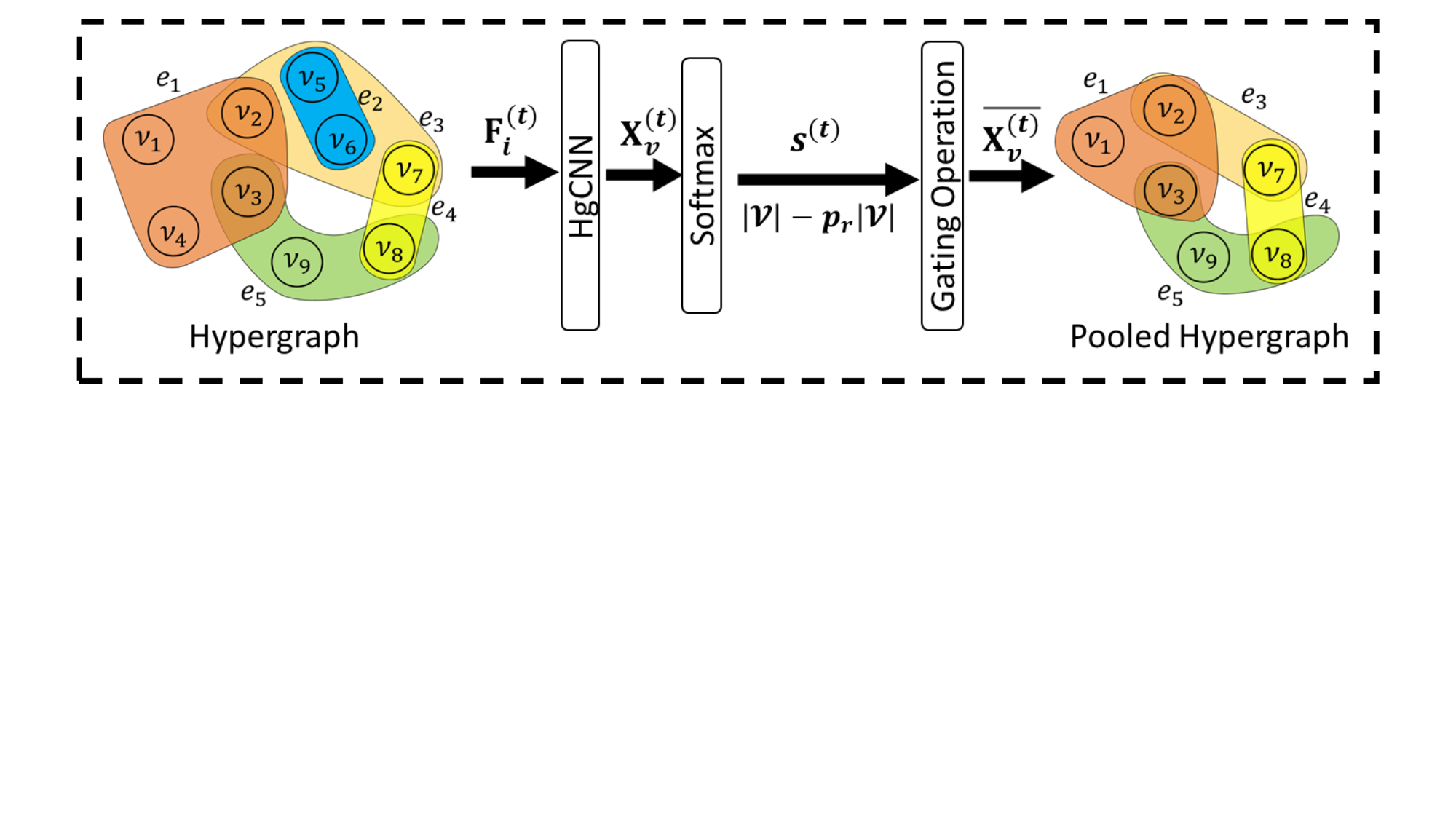}
    \vspace{-45mm}
    \caption{The HgPool operator samples a subhypergraph to learn hierarchical representations.}
    \label{fig:c(HgPool)}
\end{figure}

\vspace{-3mm}
where  $\mathbf{W}_{6}, \mathbf{W}_{7} \in \mathbb{R}^{d \times d}$, $\mathbf{W}_{8} \in \mathbb{R}^{3d}$ and $\bar{x}^{(t)}_{v_{i}} , \bar{x}^{(t)}_{e_{p}} \in \mathbb{R}^{d}$. $\kappa_{i, p}$ and $\zeta_{i, p}$ denote the attention score and coefficient. The parametric HgPool operator sequentially performs (a) the local-hypergraph pooling operations to reduce the hypergraph size, (b) applies gating operation, and (c) higher-order message passing schemes to learn hierarchical representations while preserving the prominent hypergraph structural information. Figure \ref{fig:c(HgPool)} illustrates the HgPool module.

\vspace{-3mm}

\vspace{-4mm}
\subsubsection{Hypergraph UnPooling(UnHgPool)}\label{unhgpool}
\begin{figure}[htbp]
\centering
\begin{minipage}{.35\linewidth}
\centering
\begin{tikzpicture}[scale=0.6]
\begin{scope}[yshift=-6cm]
        \node (v1) at (0, 2) {};
        \node (v3) at (4, 2.5) {};
        \node (v4) at (0, 0) {};
        \node (v5) at (2, 0.5) {};
        \node (v6) at (3.5, 0) {};

        \begin{scope}[fill opacity = 0.8]
            \filldraw [fill = yellow!70] ($(v1) + (-0.5, 0)$)
            to [out = 90, in = 180] ($(v3) + (0, 0.5)$)
            to [out = 0,in = 90] ($(v3) + (1, 0)$)
            to [out = 270, in = 0] ($(v1) + (1, -0.8)$)
            to [out = 180, in = 270] ($(v1) + (-0.5, 0)$);
            \filldraw [fill = blue!50] ($(v4) + (-0.5, 0.2)$)
            to [out = 90, in = 180] ($(v1) + (0, 1)$)
            to [out = 0, in = 90] ($(v4) + (0.6, 0.3)$)
            to [out = 270, in = 0] ($(v4) + (0, -0.6)$)
            to [out = 180, in = 270] ($(v4) + (-0.5, 0.2)$);
            \filldraw [fill = green!70] ($(v5) + (-0.5, 0)$)
            to [out = 90, in = 225] ($(v3) + (-0.5, -1)$)
            to [out = 45, in = 270] ($(v3) + (-0.7, 0)$)
            to [out = 90, in = 180] ($(v3) + (0, 0.5)$)
            to [out = 0, in = 90] ($(v3) + (0.7, 0)$)
            to [out = 270, in = 90] ($(v3) + (-0.3, -1.8)$)
            to [out = 270, in = 90] ($(v6) + (0.5, -0.3)$)
            to [out = 270, in = 270] ($(v5) + (-0.5, 0)$);
  
        \end{scope}
        \begin{pgfonlayer}{background}
        \draw[edge,color=orange] (v4) -- (v5);
        \end{pgfonlayer}

        \foreach \i in {1, 3, ..., 5}
        {
            \fill (v\i) circle (0.1);
        }

        \fill (v1) circle (0.1) node [right] {$v_1$};
        \fill (v3) circle (0.1) node [left] {$v_3$};
        \fill (v4) circle (0.1) node [below] {$v_4$};
        \fill (v5) circle (0.1) node [below right] {$v_5$};
\end{scope}               
\end{tikzpicture}
\end{minipage}%
\hspace{-3.5cm}\begin{minipage}{.5\linewidth}
\centering
\begin{tikzpicture}
  \coordinate (A) at (0,0);
  \coordinate (B) at (3,0);
  \draw[->] (A) -- (B) node[midway,fill=white] {\text{UnHgPool}};
\end{tikzpicture}
\end{minipage}%
\hspace{-3.5cm}\begin{minipage}{.45\linewidth}
\centering
\begin{tikzpicture}[scale=0.6]
\begin{scope}[yshift=-6cm]
        \node (v1) at (0, 2) {};
        \node (v2) at (1.5, 3) {};
        \node (v3) at (4, 2.5) {};
        \node (v4) at (0, 0) {};
        \node (v5) at (2, 0.5) {};
        \node (v6) at (3.5, 0) {};

        \begin{scope}[fill opacity = 0.8]
            \filldraw [fill = yellow!70] ($(v1) + (-0.5, 0)$)
            to [out = 90, in = 180] ($(v2) + (0, 0.5)$)
            to [out = 0,in = 90] ($(v3) + (1, 0)$)
            to [out = 270, in = 0] ($(v2) + (1, -0.8)$)
            to [out = 180, in = 270] ($(v1) + (-0.5, 0)$);
            \filldraw [fill = blue!50] ($(v4) + (-0.5, 0.2)$)
            to [out = 90, in = 180] ($(v1) + (0, 1)$)
            to [out = 0, in = 90] ($(v4) + (0.6, 0.3)$)
            to [out = 270, in = 0] ($(v4) + (0, -0.6)$)
            to [out = 180, in = 270] ($(v4) + (-0.5, 0.2)$);
            \filldraw [fill = green!70] ($(v5) + (-0.5, 0)$)
            to [out = 90, in = 225] ($(v3) + (-0.5, -1)$)
            to [out = 45, in = 270] ($(v3) + (-0.7, 0)$)
            to [out = 90, in = 180] ($(v3) + (0, 0.5)$)
            to [out = 0, in = 90] ($(v3) + (0.7, 0)$)
            to [out = 270, in = 90] ($(v3) + (-0.3, -1.8)$)
            to [out = 270, in = 90] ($(v6) + (0.5, -0.3)$)
            to [out = 270, in = 270] ($(v5) + (-0.5, 0)$);
            \filldraw [fill = red!70] ($(v2) + (-0.5, -0.2)$)
            to [out = 90, in = 180] ($(v2) + (0.2, 0.4)$)
            to [out = 0, in = 180] ($(v3) + (0, 0.3)$)
            to [out = 0, in = 90] ($(v3) + (0.3, -0.1)$)
            to [out = 270, in = 0] ($(v3) + (0, -0.3)$)
            to [out = 180, in = 0] ($(v3) + (-1.3, 0)$)
            to [out = 180, in = 270] ($(v2) + (-0.5, -0.2)$);
            \filldraw [fill = purple!50] ($(v5) + (-0.5, -0.2)$)
            to [out = 90, in = 180] ($(v5) + (0.2, 0.4)$)
            to [out = 0, in = 180] ($(v6) + (0, 0.3)$)
            to [out = 0, in = 90] ($(v6) + (0.3, -0.1)$)
            to [out = 270, in = 0] ($(v6) + (0, -0.3)$)
            to [out = 180, in = 0] ($(v6) + (-1.3, 0)$)
            to [out = 180, in = 270] ($(v5) + (-0.5, -0.2)$);         
        \end{scope}
        \begin{pgfonlayer}{background}
        \draw[edge,color=orange] (v4) -- (v5);
        \end{pgfonlayer}

        \foreach \i in {1, 2, ..., 6}
        {
            \fill (v\i) circle (0.1);
        }

        \fill (v1) circle (0.1) node [right] {$v_1$};
        \fill (v2) circle (0.1) node [below right] {$v_2$};
        \fill (v3) circle (0.1) node [left] {$v_3$};
        \fill (v4) circle (0.1) node [below] {$v_4$};
        \fill (v5) circle (0.1) node [below right] {$v_5$};
        \fill (v6) circle (0.1) node [below left] {$v_6$};
\end{scope}               
\end{tikzpicture}
\end{minipage}\par\medskip
\vspace{-6mm}
\caption{The UnHgPool operator performs the upsampling operation. For illustration purposes, we add the formerly rejected hypernodes($v_2, v_6$) and the corresponding hyperedges($e_2, e_6$) to the pooled-hypergraph(left) to obtain the hypergraph(right).}
\label{fig:d}
\end{figure}
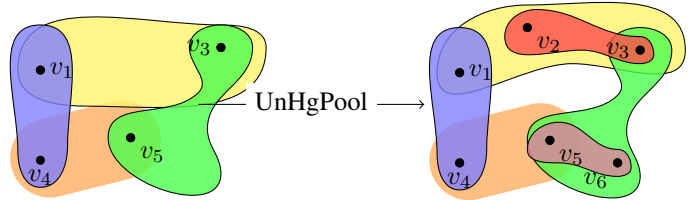

\vspace{-2mm}
In contrast, the hypergraph unpooling operator performs the inverse operation for the local- and global-neighborhood enlargement by upsampling the pooled hypergraph $\mathcal{\bar{G}}^{(t)}$ to the original hypergraph structure $\mathcal{G}^{(t)}$. The UnHgPool operator utilizes the indices of the hypernodes selected by the HgPool operator to restore the high-resolution hypergraph topology. Figure \ref{fig:d} illustrates the upsampling operation on the pooled hypergraph. The forward propagation of the hypergraph unpooling mechanism given by,

\vspace{-6mm}
\begin{align}
\mathbf{X}^{(t)}_{v} &= \mbox{addition}(0^{(t)}_{n \times d}, \mathbf{\bar{X}}^{(t)}_{v}, \mbox{idx}^{(t)}) \label{eq:unhgpool5} 
\end{align} 

\vspace{-3mm}
The vector $\mbox{idx}^{(t)}$ contains the indices of the important hypernodes selected from the hypernode-ranking operation of the Hgpool operator. $\mathbf{\bar{X}}^{(t)}_{v} \in \mathbb{R}^{n_p \times d}$ denotes the hypernode attribute matrix of the pooled hypergraph obtained from Equation \eqref{eq:hgpool5}. $0^{(t)}_{n \times d} \in \mathbb{R}^{n \times d}$ denotes the initial hypernode attribute matrix of the resultant upsampled hypergraph from the UnHgPool operator. The mathematical operation, $\mbox{addition}(0^{(t)}_{n \times d}, \mathbf{\bar{X}}^{(t)}_{v}, \mbox{idx})$, distributes the hypernode representations from the $\mathbf{\bar{X}}^{(t)}_{v}$ matrix into the $0^{(t)}_{n \times d}$ matrix, according to the indices stored in $\mbox{idx}$. In simple terms, row vectors in $0^{(t)}_{n \times d}$ are replaced with the corresponding row vectors in $\mathbf{\bar{X}}^{(t)}_{v}$, acceding with indices in $\mbox{idx}$ and the rest of the row vectors filled with zeros. Furthermore, we have skip connections to perform the feature-map summation between the hypernode representations obtained from the HgPool operator(refer to Equation \ref{eq:hgpool1}) and the upsampled hypergraph(refer to Equation \ref{eq:unhgpool5}). In brief, we fill the zero vectors in the hypernode attribute matrix with the corresponding row vectors obtained by the HgPool operator. We then compute the hyperedge attribute matrix $\mathbf{X}^{(t)}_{e}$ of the upsampled hypergraph as given by,
 
\vspace{-5mm}
\begin{gather}
x^{(t)}_{e_{p}} = \sigma \big( \hspace{-1mm}\sum_{i \hspace{0.5mm}\in \hspace{0.5mm}{\mathcal{N}_{p}}} \hspace{-1mm} \mathbf{W}_{9} x^{(t)}_{v_{i}}\big)  \nonumber 
\end{gather}
 
\vspace{-3mm} 
where $\mathbf{W}_{9} \in \mathbb{R}^{d \times d}$. Further, we employ the self-attention mechanism to explicitly model the anomaly propagation delay or advancement from the global-neighborhood sooner or later through the hypergraph topology. The refined hypernode-level representations are obtained by,

\vspace{-4mm}
\begin{equation}
\mathbf{x}^{(t)}_{v_{i}} = \mathbf{W}_{9}\mathbf{x}^{(t)}_{v_{i}} + \sum_{p \in \mathcal{N}_{i}} \gamma_{i,p}  \mathbf{W}_{10} \mathbf{x}^{(t)}_{e_{p}}  \label{eq:unhgpool1}
\end{equation}

\vspace{-3mm}
where $\mathbf{W}_{10} \in \mathbb{R}^{d \times d}$ and $\mathbf{W}_{10} \mathbf{x}^{(t)}_{e_{p}}$ denotes the information propagated from the hyperedge $p$ to the hypernode $i$. The attention score($\delta_{i,p}$) and coefficient($\gamma_{i,p}$) are evaluated by,

\vspace{-5mm}
\begin{align}
\mathbf{g}_{i}^{(t)} &= \mathbf{W}_{11}(\mathbf{z}_{i} \oplus \mathbf{W}_{9} \mathbf{x}^{(t)}_{v_{i}}); \mathbf{g}_{p}^{(t)} = \mathbf{W}_{10} \mathbf{x}^{(t)}_{e_{p}} \label{eq:unhgpool2} \\
\delta_{i,p} &= \big(\mathbf{W}_{12}\big(\mathbf{g}_{p}^{(t)}-\mathbf{g}_{i}^{(t)} \big)\big)^{\top} \mathbf{W}_{12}\big( \mathbf{g}_{p}^{(t)} + \mathbf{g}_{i}^{(t)}\big) \label{eq:unhgpool3}  
\end{align} 

\vspace{-6mm}
\begin{align}
\gamma_{i, p} &= \frac{\exp \left(\delta_{i,p}\right)}{\sum_{k \in \mathcal{N}_{i}} \exp \left(\delta_{i,k}\right)} \label{eq:unhgpool4}
\end{align} 

\vspace{-1mm}
where $\mathbf{W}_{11} \in \mathbb{R}^{d \times 2d}$, $\mathbf{W}_{12} \in \mathbb{R}^{d \times d}$ are learnable weight matrices. In a nutshell, the UnHgPool operator upsamples the pooled hypergraph and learns the expressive hierarchical hypernode representations through the self-attention mechanism-based neighborhood aggregation schemes.  

\vspace{-6mm}

\vspace{-5mm}
\begin{figure}[htbp]
\centering
\begin{tikzpicture}[auto, node distance=2cm,>=latex']
    \node [input, name=input] {};
    \node [sum, right of=input] (sum) {};
    \node [block, thick, right of=sum] (controller) {$\text{HgPool}$};
    \node [block, thick, right of=controller, node distance=3cm] (system) {$\text{UnHgPool}$};
    
    \draw [->] (sum) -- node {$\mathcal{G}^{(t)}$} (controller);
    \node [output, right of=system] (output) {};
     
    \draw [->] (controller) -- node[label={[xshift=0.0cm, yshift=-0.2cm]$\mathcal{\bar{G}}^{(t)}$}] {} (system);
    \draw [->] (system) -- node[label={[xshift=0cm, yshift=-0.2cm]$ $}] {$\mathcal{G}^{(t)}$} (output);
    
\end{tikzpicture}
\vspace{-1mm}
\caption{Overview of the $\text{HgED}$ module.} \label{fig:e}
\end{figure}
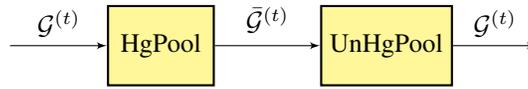

\vspace{-3mm}
Figure \ref{fig:e} illustrates the $\text{HgED}$ module. The module proposes a single-stage hierarchical encoder-decoder method to learn the hierarchical representations from fine-to-coarse and coarse-to-fine transformation of the hypergraph. The hypergraph encoder had modeled by the HgPool operator. The encoder downsamples the original hypergraph structure by local-hypergraph pooling operations, thereby transforming the hypernode representations by performing information diffusion through the high-order neighborhood-aggregation schema by exploiting the pooled hypergraph structure. The hypergraph decoder had modeled by the UnHgpool operator. The decoder upsamples the pooled hypergraph and refines the hypernode representations through the self-attention mechanism-based neighborhood filtering operations. In essence, the hypergraph encoder-decoder module learns the prominent subhypergraph patterns to model the long-range spatial relations and long-term temporal dependencies of the hypergraph structured data.

\vspace{-3mm}
\subsection{Hypergraph-Forecasting(HgF)}\label{hgf}
\vspace{-3mm}
The hypergraph forecasting module predicts the multi-sensor values at time step $t$, i.e., $\hat{\mathbf{f}}^{(\mathbf{t})} \in \mathbb{R}^{n}$. 

\vspace{-4mm}
\begin{equation}
\hat{\mathbf{f}}^{(\mathbf{t})} = f_{\theta}\big(\big[\mathbf{z}_{1} \odot \mathbf{x}^{(t)}_{v_{1}} \oplus \cdots \oplus \mathbf{z}_{n} \odot \mathbf{x}^{(t)}_{v_{n}}\big]\big)
\end{equation}

\vspace{-2mm}
where the hypernode representations($\mathbf{x}^{(t)}_{v_{i}}$) obtained from HgED. Here, the goal is to minimize the mean squared error(MSE) between the model predictions $\hat{\mathbf{f}}^{(\mathbf{t})}$ and the observed data $\mathbf{f}^{(\mathbf{t})}$ for lower-forecasting error. Under a scenario where labeled-anomaly data are available. An anomaly detection task is a supervised two-category classification, where the objective is to distinguish the normal and abnormal instances. The hypergraph prediction module in this scenario,

\vspace{-4mm}
\begin{equation}
\hat{\mathbf{y}}^{(\mathbf{t})} = \sigma f_{\gamma}\big(\big[\mathbf{z}_{1} \odot \mathbf{x}^{(t)}_{v_{1}} \oplus \cdots \oplus \mathbf{z}_{n} \odot \mathbf{x}^{(t)}_{v_{n}}\big]\big)
\end{equation}

\vspace{-2mm}
where the goal is to minimize the binary cross-entropy loss between the model predictions($\hat{\mathbf{y}}^{(\mathbf{t})} \in {0,1}$) and the ground-truth labels($\mathbf{y}^{(\mathbf{t})} \in {0,1} $). $f_{\theta}, f_{\gamma}$ are fully-connected layers. 

\vspace{-2mm}
\subsection{Hypergraph-Deviation(HgD)}\label{hgd}
\vspace{-2mm}
The HgD module in the unsupervised anomaly detection task computes the robust normalized anomaly scores($A^{(t)}_{i}$). This information regarding the sensors help in accurately localizing the anomalies within the multisensor data in the temporal domain.

\vspace{-5mm}
\begin{equation}
A^{(t)}_{i}=\frac{\operatorname{dev}^{(t)}_{i}-\tilde{\mu}_{i}}{\tilde{\sigma}_{i}}; \operatorname{dev}^{(t)}_{i}=\left|\mathbf{f}^{(\mathbf{t})}_{i}-\hat{\mathbf{f}}^{(\mathbf{t})}_{i}\right|
\end{equation}

\vspace{-2.5mm}
where $\hat{\mathbf{f}}^{(\mathbf{t})}_{i}$, $\mathbf{f}^{(\mathbf{t})}_{i}$ are prediction of HgF module and ground truth respectively. ${\tilde{\mu}_{i}}$ and $\tilde{\sigma}_{i}$ are  the  median(second quartile) and inter-quartile range(difference of upper and lower quartiles) of $A^{(t)}_{i}$ across the time points. We compute the simple moving average of the maximum value of anomalousness score($A^{(t)}_{i}$) across the multi sensors at time point $t$ over the validation set as given,

\vspace{-7mm}
\begin{gather}
\text{Th} =  \underset{t \in  \mathcal{T}_{val}}{\max}A^{(t)} ; A^{(t)} = \frac{1}{w_{a}} \sum_{t-(w_{a}+1)}^{t} \underset{i \in  \mathcal{V}}{\max} \big(A^{(t)}_{i} \big) \nonumber
\end{gather}

\vspace{-4mm}
where $w_{a}(=10)$ denotes the number of time points in the moving average calculation.  $\mathcal{T}_{val}$ denotes the time points in the validation set. We set the anomaly detection threshold($\text{Th}$) as the max of $A^{(t)}$ over the validation data. During inference, time points in the test set with an anomaly score higher than the threshold score were identified as abnormality events. 

\vspace{-5mm}
\section{Experiments and results}\label{sec:experiments}
\vspace{-3mm}
We demonstrate and support the effectiveness of our proposed method in comparison to the baseline approaches by investigating the following research questions:

\vspace{-2mm}
\begin{itemize}
\item RQ1(Accuracy): How does our proposed method perform compared to the baseline methods on the anomaly-detection task$?$
\item RQ2(Ablation): How helpful are the modules of the proposed method in improving the model performance?
\item RQ3(Interpretability): How does our method provide the anomaly diagnosis and suggest recommendations in a prescriptive approach to avoid the anomalies?
\end{itemize}

\vspace{-5mm}
\subsection{Benchmark Datasets}\label{sec:datasets}    
\vspace{-3mm}
To probe the efficacy of our proposed method on anomaly detection, we evaluate our framework on a variety of publicly available datasets for competitive benchmarking with the baseline methods. Table~\ref{table:dataset} summarizes the characteristics of the varied datasets used in this study. The \text{SWaT} and \text{WADI}\footnote{{https://itrust.sutd.edu.sg/itrust-labs\_datasets/}} are real-world datasets on water treatment facilities and distribution networks. \text{SMAP} and \text{MSL} are expert-labeled open-sourced datasets of telemetry data from NASA\cite{hundman2018detecting}. The Tennessee Eastman process(TEP)\footnote{https://dataverse.harvard.edu/dataverse/harvard} is a simulated industrial benchmark dataset for process monitoring and control. It contains 20 different faults.
The HAI\footnote{https://github.com/icsdataset/hai} is a time-series dataset of a realistic industrial testbed of steam-turbine power generation and pumped-storage hydropower generation. It contains 38 different attack scenarios.  

\vspace{-5mm}
\begin{table}[htbp] 
 \caption{Statistical summary of benchmarked datasets.}
 \vspace{1mm}
  \label{table:dataset}
  \centering
  \resizebox{0.6\linewidth}{!}{%
  \begin{tabular}{lrrrrcr}
    \toprule
    Dataset & SWaT & WADI & SMAP & MSL & TEP & HAI\\
    \midrule
    sensors(\textit{n}) & 51 & 123 & 25 & 55 & 52 & 59\\
    \textit{w} & 30 & 30 & 60 & 60 & 30 & 25\\
    \textit{k} & 15 & 25 & 7 & 10 & 20 & 20\\
   \bottomrule 
\end{tabular}
}
\end{table}

\vspace{-3mm}
\subsection{Experimental setup}\label{sec:setup}
\vspace{-3mm}
We perform min-max scaling on all the datasets to rescale variables into the range [0,1]. We train the model for 100 epochs with a batch size of 48 on multiple NVIDIA Tesla T4 GPUs in all the experiments.  We optimize using Adam optimizer with initial learning rate($\text{lr}$) set as 0.001 and $(\beta1,\beta2)$ =(0.9,0.99). We implement the early-stopping technique and adopt the $\text{lr}$ adjusting strategy to decay the $\text{learning rate}$ by half with the patience of 10 epochs on the validation set. We conduct five different experimental runs and report the mean values of the evaluation metrics obtained on the different experimental run outputs across all the datasets.

\subsection{Model configurations}\label{sec:modelconfig}
\vspace{-3mm}
The hypernode embeddings($z_{i}$) and representations($x^{(t)}_{v_{i}}$) dimensions have a fixed size($d$) of 128 for all the datasets. We report, in Table~\ref{table:dataset}, the optimal sliding window size($w$) and the number of nearest neighbors($k$) used in this study across the datasets.

\vspace{-4mm}
\subsection{Evaluation Metrics}
\vspace{-2mm}
We evaluate and report the performance of our model on the test set across all the benchmark datasets. We report the model performance in terms of the standard evaluation metrics such as precision(P as $\%$), recall(R as $\%$), and F1-score(F1 as $\%$) for a fair and rigorous comparison with the baseline models. We utilize the maximum anomaly score(refer to section \ref{hgd}) over the validation dataset to set the threshold to identify anomalies. For SWaT and WADI datasets, there exist contiguous anomaly segments. We had adopted the point adjustment strategy\cite{shen2020timeseries, zhao2020multivariate} to flag the entire time segment as an anomaly if the model predictions within this subset window had detected an anomaly.

\vspace{-4mm}
\subsection{Results} 
\vspace{-2mm}
We compare the $\textbf{HgAD}$ model with state-of-the-art methods for multivariate time-series anomaly detection on all the datasets.  Tables \ref{tab:results} and \ref{tab:TES} present the performance of the baseline models in comparison with our proposed method. The results show that the $\textbf{HgAD}$ model outperforms the baseline models and attains significant gains consistently across the evaluation metrics on all datasets. On SWaT and WADI datasets, the proposed method demonstrates relatively the best performance with a high precision score of $97.48\%$ on SWaT and $98.75\%$ on WADI. The HgAD model brings a moderate $2\%$  and $1.3\%$ improvement in the precision score compared to the GRELEN\cite{zhanggrelen} and GDN\cite{deng2021graph} on SWaT and WADI datasets, respectively. In terms of recall and F1-measure, the $\textbf{HgAD}$ model surpasses the previous state-of-the-art baselines and reports a percentage increase of $2.2\%$ and $3.9\%$ on SWaT; $1.7\%$, $10.65\%$ on WADI over the successive-best baseline models GTA\cite{chen2021learning}, GRELEN\cite{zhanggrelen} and GTA\cite{chen2021learning}, MTAD-GAT\cite{zhao2020multivariate} respectively. On SMAP and MSL datasets, the $\textbf{HgAD}$ model unsurprisingly outperforms the baseline models and reports the highest scores on all the evaluation metrics. In terms of precision score, the $\textbf{HgAD}$ model shows a phenomenal increment of $3.7\%$ on SMAP; $4\%$ on MSL compared to the next-best baseline model GRELEN\cite{zhanggrelen}. Similarly, our method attains the best performance and reports a high recall and F1 score of $99.07\%$ and $99.23\%$ on SMAP; $97.11\%$ and $96.84\%$ on MSL, respectively. Further, in the multi-fault anomaly detection task on the Tennessee Eastman simulated datasets, the $\textbf{HgAD}$ model surpasses the baseline models on fault detection rate(FDR, \cite{fu2022mad}). In particular, the $\textbf{HgAD}$ model outperforms the next-best baseline models by a large margin on the identification of faults 3, 5, 9, and 10 with a remarkable improvement of $31.2\%$, $23.81\%$, $29.07\%$, and $27.09\%$, respectively. Likewise, on faults 15, 16 and, 19, the $\textbf{HgAD}$ model shows an increment of $19.6\%$, $23.5\%$, $21.89\%$ over the succeeding-best baseline models, respectively. In addition, on the HAI dataset, the $\textbf{HgAD}$ model reports relative improvements of $18.1\%$ in the precision score, $14.4\%$ in the recall, and $15.3\%$ in the F1 score. In brief, our method showed impressive improvements and demonstrated its potential to detect anomalous events on unbalanced and high-dimensional datasets of great relevance to industry. We generated and verified the results of the baseline models from \cite{xu2021anomaly, deng2021graph, chen2021learning, fu2022mad}.

\vspace{-4mm}
\subsection{Ablation studies}
\vspace{-2mm}
\label{sec:ablation}  
We perform ablation studies to provide insights into the relative contribution of the different architectural modules of our proposed method for the improved overall performance of the $\textbf{HgAD}$ model. We gradually exclude  or substitute the modules to design several variants of our proposed method. We examine the variant's performance compared to the $\textbf{HgAD}$ model on all the datasets. Table \ref{tab:ablation} shows the performance comparison results of the $\textbf{HgAD}$ model and its variants in the ablation studies. (a) We study the efficacy of the hypergraph structure learning module in modeling the complex relational dependencies among the multiple IoT sensors in the multisensor data. We utilize the HgSL module to construct the k-uniform nearest neighbor hypergraph (k-uniform NNHg) representation of the multisensor data. Please refer to subsection\ref{hgsl}. We refer to the $\textbf{HgAD}$ model realized with the following limiting scenarios of structure modeling as follows, 

\vspace{-1.5mm}
\begin{itemize}
\item $\textbf{w/} \hspace{1mm} \textbf{2-graph}$: \hspace{-1mm} $\textbf{HgAD}$ model with 2-uniform NNHG(the hyperedges have the same cardinality of size 2).
\item $\textbf{w/} \hspace{1mm} \textbf{irr-hypergraph}$: $\textbf{HgAD}$ model with irregular hypergraph(hyperedges have different cardinality) computed through the Gumbel-softmax sampling technique\cite{chen2021learning}.
\end{itemize}

\vspace{-1mm}
As shown in Table \ref{tab:ablation}, we notice a drop in the performance of the variants compared to our proposed method. The $\textbf{HgAD}$ model significantly outperforms the variants($\textbf{w/} \hspace{1mm} \textbf{2-graph}$, $\textbf{w/} \hspace{1mm} \textbf{irr-hypergraph}$), corroborating the appeal of learning the optimal discrete hypergraph structure with nearest neighbors search. In terms of precision measure, we observe an $18.6\%$ and $4.2\%$ drop in SWaT and; $13.1\%$ and $2.1\%$ drop in WADI of the prior mentioned variants compared to the $\textbf{HgAD}$ model. The results show the advantages of utilizing the HgSL module to capture the underlying structural information in a hypergraph representation of the observed multisensor data. (b) We study the impact of hypernode positional embeddings. Please refer to subsection\ref{hpe}. We refer to the $\textbf{HgAD}$ model in the absence or substitute with an alternative positional encoding(PE) technique as follows: 

\vspace{-0.5mm}
\begin{itemize}
\item $\textbf{w/o PosEmb}$: $\textbf{HgAD}$ model without the PEs.
\item $\textbf{w/ SPEs}$: $\textbf{HgAD}$ model with Sinusoidal PEs.
\end{itemize}

\vspace{-1mm}
The $\textbf{HgAD}$ model performs better than both variants, as illustrated in Table \ref{tab:ablation}. The precision scores of the design variant, $\textbf{w/o PosEmb}$ declined by $3.35\%$ on SWaT;  $5.65\%$ on WADI compared to the $\textbf{HgAD}$ model. The impact of $\textbf{w/ SPEs}$ is marginal and achieves on-par performance compared to the $\textbf{HgAD}$ model. The results justify the purpose for the inclusion of the positional encoding scheme and support our approach of learning the optimal position embeddings through the training parameters of the model. 

\begin{table*}[htbp]
\center
\setlength{\tabcolsep}{5pt}
\caption{Experimental results on the anomaly detection benchmark datasets in terms of precision, recall, and F1-score}
\vspace{-2mm}
\label{tab:results}
\resizebox{1.1\linewidth}{!}{%
\hspace{-12mm}\begin{tabular}{@{}c|ccc|lll|ccc|ccc|ccc@{}}
\toprule
\multirow{2}{*}{\textbf{Methods}} & \multicolumn{3}{c|}{\textbf{SWaT}}                                  & \multicolumn{3}{c|}{\textbf{WADI}}                                                                         & \multicolumn{3}{c|}{\textbf{SMAP}}                                  & \multicolumn{3}{c}{\textbf{MSL}}    & \multicolumn{3}{c}{\textbf{HAI}}                               \\ \cmidrule(l){2-4} \cmidrule(l){5-7} \cmidrule(l){8-10} \cmidrule(l){11-13} \cmidrule(l){14-16}
                                  & \textbf{P(\%)}       & \textbf{R(\%)}       & \textbf{F1(\%)}          & \multicolumn{1}{c}{\textbf{P(\%)}} & \multicolumn{1}{c}{\textbf{R(\%)}} & \multicolumn{1}{c|}{\textbf{F1}} & \textbf{P(\%)}       & \textbf{R(\%)}       & \textbf{F1(\%)}          & \textbf{P(\%)}       & \textbf{R(\%)}       & \textbf{F1(\%)}     & \textbf{P(\%)}       & \textbf{R(\%)}       & \textbf{F1(\%)}      \\ \midrule
PCA\cite{shyu2003novel}                               & 24.92                & 21.63                & 23.0                 & \multicolumn{1}{c}{39.53}          & \multicolumn{1}{c}{5.63}           & \multicolumn{1}{c|}{10.0}        &           33.62           &    29.56                  &       26.72               &                     25.07 &  31.56                    &    28.44     &  10.73                 & 9.31                 & 7.35            \\
KNN\cite{angiulli2002fast}                               & 11.29                 & 9.47                 & 8.75                 & \multicolumn{1}{c}{7.76}           & \multicolumn{1}{c}{7.75}           & \multicolumn{1}{c|}{8.0}        &  10.47                    &        9.84              &          8.96            &                     12.43 &          10.28            &        9.83           &  12.53                 & 9.82                 & 8.35   \\
KitNet\cite{mirsky2018kitsune}                            & \multicolumn{1}{l}{79.11} & \multicolumn{1}{l}{83.17} & \multicolumn{1}{l|}{76.37} &                                   83.61 &             81.78                       &          80.09                       & 77.25                & 83.27                & 80.14                 & 63.12                & 79.36                & 70.31               &  27.56                 & 24.18                 & 21.86  \\
GAN-Li\cite{li2018anomaly}                            & \multicolumn{1}{l}{81.03} & \multicolumn{1}{l}{84.97} & \multicolumn{1}{l|}{77.32} &                                   76.25 &               80.33                     &      77.95                           & 67.10                & 87.06                & 75.19                 & 71.02                & 87.06                & 78.23              &  19.83                 & 18.36                 & 17.45   \\
LSTM-NDT\cite{hundman2018detecting}                          & \multicolumn{1}{l}{79.12} & \multicolumn{1}{l}{75.08} & \multicolumn{1}{l|}{78.75} &                                   81.25 &                       78.64             &      75.18                           & 89.65                & 88.46                & 89.05                 & 59.44                & 53.74                & 56.40                &  22.46                 & 23.45                 & 20.32 \\
MTAD-GAT\cite{zhao2020multivariate}                          & \multicolumn{1}{l}{82.01} & \multicolumn{1}{l}{76.84} & \multicolumn{1}{l|}{72.47} &                                   82.58 &                  84.94                  &      \underline{80.25}                           & 89.06                & 91.23                & 90.41                 & 87.54                & \underline{94.40}                & 90.84       &  24.75                 & 21.78                 & 20.14            \\
FB\cite{lazarevic2005feature}                                & 10.17                & 10.17                & 10.0                 & \multicolumn{1}{c}{8.60}           & \multicolumn{1}{c}{8.60}           & \multicolumn{1}{c|}{9.0}        &     16.47                 &         14.37             &       12.95               &                     14.65 &      11.73               &    12.38                 &  8.11                 & 8.92                 & 8.54   \\
AE\cite{aggarwal2017introduction}                                & 72.63                & 52.63                & 61.0                 & \multicolumn{1}{c}{34.35}          & \multicolumn{1}{c}{34.35}          & \multicolumn{1}{c|}{34.0}        &           59.76            &         54.74              &     51.35                  &                      57.21  &            56.67           &       53.07              &  15.03                 & 17.34                 & 14.67  \\
MAD-GAN\cite{li2019mad}                           & \underline{98.97}                & 63.74                & 77.0                 & \multicolumn{1}{c}{41.44}          & \multicolumn{1}{c}{33.92}          & \multicolumn{1}{c|}{37.0}        & 80.49                & 82.14                & 81.31                 & 85.17                & 89.91                & 87.47            &  25.27                 & 23.34                 & 21.87    \\
GDN\cite{deng2021graph}                               & \textbf{99.35}                & 68.12                & 81.0                 & \multicolumn{1}{c}{\underline{97.50}}          & \multicolumn{1}{c}{40.19}          & \multicolumn{1}{c|}{57.0}        & 86.62                & 84.27                & 83.24                 & 89.92               & 87.24                & 86.84           &  43.41                 & 46.27                 & \underline{44.59}     \\
GTA\cite{chen2021learning}                         & 74.91                &  \underline{96.41}               & 84.0                 & \multicolumn{1}{c}{74.56}          & \multicolumn{1}{c}{\underline{90.50}}          & \multicolumn{1}{c|}{82.0}        & 89.11                & 91.76                & 90.41                 & 91.04                & 91.17                & 91.11          &  44.91                 & 41.63                 & 40.29        \\ 
OCSVM\cite{scholkopf2001estimating}                             & 45.39                & 49.22                & 47.23                &                                   49.17  &        53.23                            &    51.91                             & 53.85                & 59.07                & 56.34                 & 59.78                & 86.87                & 70.82              &  35.71                 & 33.08                 & 31.19   \\
IsolationForest\cite{liu2008isolation}                   & 49.29                & 44.95                & 47.02                &              52.37  &        56.41 &                55.06                                                      & 52.39                & 59.07                & 55.53                 & 53.94                & 86.54                & 66.45  &  33.09                 & 35.49                 & 30.77\\
LOF\cite{breunig2000lof}                               & 72.15                & 65.43                & 68.62                &      57.02                              &                                   61.17 &     53.46                            & 58.93                & 56.33                & 57.60                 & 47.72                & 85.25                & 61.18       &  31.27                 & 29.93                 & 26.48\\
Deep-SVDD\cite{ruff2018deep}                         & 80.42                & 84.45                & 82.39                &                74.18                    &                                   70.82 &              73.43                   & 89.93                & 56.02                & 69.04                 & 91.92                & 76.63                & 83.58                 &  34.81                 & 31.26                 & 30.94\\
DAGMM\cite{zong2018deep}                            & 89.92                & 57.84                & 70.4                  & \multicolumn{1}{c}{54.44}          & \multicolumn{1}{c}{26.99}          & \multicolumn{1}{c|}{36.0}        & 86.45                & 56.73                & 68.51                 & 89.60                & 63.93                & 74.62          &  35.56                 & 37.12                 & 33.77       \\  
MMPCACD\cite{yairi2017data}                           & 82.52                & 68.29                & 74.73                &         74.29                           &                                   75.01 &                     71.48            & 88.61                & 75.84                & 81.73                 & 81.42                & 61.31                & 69.95                 &  31.58                 & 29.46                 & 27.33 \\
VAR\cite{anderson1976time}                               & 81.59                & 60.29                & 69.34                &       75.59                             &                                   69.36 &     66.21                            & 81.38                & 53.88                & 64.83                 & 74.68                & 81.42                & 77.90 &  34.42                 & 36.28                 & 31.97\\
LSTM\cite{hundman2018detecting}                              & 86.15                & 83.27                & 84.69                &    68.73                                &                 62.47                   &       65.74                          & 89.41                & 78.13                & 83.39                 & 85.45                & 82.50                & 83.95              &  35.61                 & 32.84                 & 31.92   \\
CL-MPPCA\cite{tariq2019detecting}                          & 76.78                & 81.50                & 79.07                &   69.72                                 &    65.23                                &     67.32                            & 86.13                & 63.16                & 72.88                 & 73.71                & 88.54                & 80.44                 &  33.82                 & 31.74                 & 30.05 \\
ITAD\cite{shin2020itad}                              & 63.13                & 52.08                & 57.08                &     71.95                               &                                   69.39 &       65.76                          & 82.42                & 66.89                & 73.85                 & 69.44                & 84.09                & 76.07                 &  36.72                 & 33.42               & 32.47\\
LSTM-VAE\cite{park2018multimodal}                           & 76.00                & 89.50                & 82.20                 & \multicolumn{1}{c}{87.79}          & \multicolumn{1}{c}{14.45}          & \multicolumn{1}{c|}{25.0}        & 92.20                & 67.75                & 78.10                 & 85.49                & 79.94                & 82.62       &  38.25                 & 37.94               & 35.04         \\
BeatGAN\cite{zhou2019beatgan}                           & 64.01                & 87.46                & 73.92                &    74.46                                &                                   70.71 &     76.52                            & 92.38                & 55.85                & 69.61                 & 89.75                & 85.42                & 87.53 &  39.41                 & 38.03               & 35.47\\
OmniAnomaly\cite{li2019enhancing}                       & 81.42                & 84.30                & 82.83                &        78.18                            &                                   80.13 &         77.24                        & 92.49                & 81.99                & 86.92                 & 89.02                & 86.37                & 87.67 &  46.29                 & 43.75               & 42.73\\
InterFusion\cite{li2021multivariate}                       & 80.59                & 85.58                & 83.01                &         81.78                           &           84.37                         &    80.21                             & 89.77                & 88.52                & 89.14                 & 81.28                & 92.70                & 86.62  &  45.72                 & 43.15               & 42.55\\
THOC\cite{shen2020timeseries}                              & 83.94                & 86.36                & 85.13                &    84.24                                 &    81.32                                 &     80.09                             & 92.06                & 89.34                & 90.68                 & 88.45                & 90.97                & 89.69               &  43.72                 & \underline{45.82}               & 43.67  \\
GRELEN\cite{zhanggrelen}                                & 95.60                & 83.50                & \underline{89.10}                &      77.30                              &                                   61.30 &        68.20                         & \underline{94.45}                & \underline{98.16}                & \underline{97.29}                 & \underline{94.36}                & 94.04                & \underline{91.58}       &  \underline{47.31}                 & 43.12               & 40.58          \\ 
Proposed                          & \multicolumn{1}{l}{97.48} & \multicolumn{1}{l}{\textbf{98.60}} & \multicolumn{1}{l|}{\textbf{92.73}} &               \multicolumn{1}{l}{\textbf{98.75}}                     &       \multicolumn{1}{l}{\textbf{92.13}}                             &   \multicolumn{1}{l|}{\textbf{89.82}}                              & \multicolumn{1}{l}{\textbf{98.05}} & \multicolumn{1}{l}{\textbf{99.07}} & \multicolumn{1}{l|}{\textbf{99.23}} & \multicolumn{1}{l}{\textbf{98.26}} & \multicolumn{1}{l}{\textbf{97.11}} & \multicolumn{1}{l}{\textbf{96.84}} & \multicolumn{1}{l}{\textbf{57.79}} & \multicolumn{1}{l}{\textbf{55.32}} & \multicolumn{1}{l}{\textbf{52.89}}\\ \bottomrule
\multicolumn{8}{l}{Best performance in bold. Second-best with underlines.} \\
\end{tabular}
}
\end{table*}

\vspace{-1mm}
(c) We study the effectiveness of hypernode embeddings. Please refer to subsections \ref{hgcnn}, \ref{hgpool}, and \ref{unhgpool}. We refer to the $\textbf{HgAD}$ model without the embeddings in Equations \ref{eq:hgcnn3} of the HgCNN operator, Equation\ref{eq:hgpool7} of the local-hypergraph pooling operator, and Equation \ref{eq:unhgpool2} of the unpooling operator to compute the attention-weighted representations are as follows, 

\begin{itemize}
\item $\textbf{w/o} \hspace{1mm} \mathbf{z_{i}}$: \hspace{-1mm}$\textbf{HgAD}$ model without hypernode embeddings($z_{i}$).
\end{itemize}

The performance degradation of the variant was evident on all the datasets, as indicated in Table \ref{tab:ablation}. The design variant, $\textbf{w/o} \hspace{1mm}  \mathbf{z_{i}}$, yields an overall $16.73\%$ relative lower precision score on SWaT; $15.54\%$ on WADI compared to the $\textbf{HgAD}$ model. The embeddings enhance the performance of the $\textbf{HgAD}$ model and are indispensable for encapsulating the complex underlying relational information among the multiple IoT sensors. The ablation studies verify the effectiveness of the learned embeddings in the $\textbf{HgAD}$ model. 

\vspace{-3mm}
\begin{table*}[htbp]
\centering
\setlength{\tabcolsep}{2.5pt}
\caption{Experimental results on simulated Tennesse Eastman dataset in terms of fault detection rate(FDR($\%$))}
\vspace{-2mm}
\label{tab:TES}
\resizebox{1.2\linewidth}{!}{%
\hspace{-15mm}\begin{tabular}{@{}c|cccccccccccccccccccc|c@{}}
\toprule
\textbf{Base Model} & \textbf{1} & \textbf{2} & \textbf{3} & \textbf{4} & \textbf{5} & \textbf{6} & \textbf{7} & \textbf{8} & \textbf{9} & \textbf{10} & \textbf{11} & \textbf{12} & \textbf{13} & \textbf{14} & \textbf{15} & \textbf{16} & \textbf{17} & \textbf{18} & \textbf{19} & \textbf{20}  \\ \midrule
Transformer\cite{vaswani2017attention}         & 99.64      & 98.45      & 5.00       & 99.96      & 28.86      & 100     & 100     & 96.43      & 5.19       & 17.48       & 77.51       & 98.20       & 94.01       & 99.97       & 5.39        & 13.43       & 91.53       & 93.76       & 25.13       & 48.05               \\
TEncoder\cite{vaswani2017attention}            & 99.66      & 98.40      & 5.03       & 99.99      & 25.74      & 100     & 100     & 95.80      & 5.17       & 15.51       & 76.83       & 97.89       & 93.66       & 99.97       & 5.35        & 12.46       & \underline{96.02}       & 93.78       & 24.40       & 46.87                \\
LSTM\cite{hochreiter1997long}                & 99.74      & 97.99      & 5.12       & 99.95      & 20.63      & 100     & 100     & 93.42      & 5.21       & 21.06       & 80.43       & 95.87       & 93.21       & 99.96       & 5.37        & 16.99       & 95.46       & 93.70       & 24.10       & \underline{71.11}               \\
TCN\cite{bai2018empirical}                 & 99.61      & 97.93      & 5.12       & 100     & 26.46      & 100     & 100     & 94.68      & 5.19       & 35.57       & 80.51       & 96.63       & 93.48       & 99.97       & 5.36        & \underline{21.10}       & 96.14       & 93.90       & 23.39       & 47.92               \\
FNet\cite{lee2021fnet}                & 99.67      & 98.64      & 4.86       & 99.18      & 25.82      & 100     & 100     & 96.76      & 18.87      & 18.87       & 76.08       & 98.11       & 94.07       & 99.96       & 5.48        & 13.74       & 91.05       & 93.70       & 24.43       & 45.59              \\ 
GTA\cite{chen2021learning}                & 98.12      & \underline{99.35}      & 5.88       & 98.04      & \underline{55.82}      & 100     & 100     & 97.34      & 20.18      & 34.33       & 79.81       & 98.72       & 96.03       & 98.21       & 7.64        & 16.69       & 92.25       & 94.78       & 26.57       & 47.31       &          \\ 
GDN\cite{deng2021graph}                & \textbf{99.81}      & 99.27      & 6.72       & 99.56      & 41.07      & 100     & 100     & 95.04      & 16.46      & 41.22       & 79.57       & \textbf{99.64}       & 95.71       & 97.58       & 7.83        & 15.64       & 92.79       & \underline{95.27}       & 27.17       & 48.81       &          \\ 
MTAD-GAT\cite{zhao2020multivariate}                & \underline{99.78}      & 98.91      & 8.92       & 99.81      & 39.33      & 100     & 100     & \underline{98.57}      & \underline{20.37}      & 43.93       & \underline{82.47}       & 99.51       & \underline{96.84}       & 99.74       & \underline{10.13}        & 16.98       & 94.47       & 94.60       & \underline{30.79}       & 58.90       &          \\ 
GRELEN\cite{zhanggrelen}                & 99.67      & 98.64      & \underline{10.86}       & 99.18      & 51.82      & 100     & 100     & 96.76      & 18.87      & \underline{48.87}       & 76.08       & 98.11       & 94.07       & 99.96       & 5.48        & 13.74       & 91.05       & 93.70       & 24.43       & 62.59       &          \\ 
Proposed            & 99.72      & \textbf{99.82}      & \textbf{15.79}       & \textbf{100}      & \textbf{73.27}      & \textbf{100}     & \textbf{100}     & \textbf{99.32}      & \textbf{28.72}      & \textbf{67.03}       & \textbf{86.84}       & \underline{99.38}       & \textbf{98.56}       & 99.63       & \textbf{12.60}        & \textbf{27.61}       & \textbf{98.64}       & \textbf{97.72}       & \textbf{39.42}       & \textbf{77.53}       &          \\ \bottomrule
\multicolumn{8}{l}{Best performance in bold. Second-best with underlines.} \\
\end{tabular}
}
\end{table*}

\vspace{0mm}
\begin{table*}[htbp]
\centering
\setlength{\tabcolsep}{5pt}
\caption{Ablation studies of $\text{HgAD}$ model and its variants on the anomaly detection task.}
\vspace{-2mm}
\label{tab:ablation}
\resizebox{1.025\linewidth}{!}{%
\hspace{-10mm}\begin{tabular}{@{}c|ccc|lll|ccc|ccc@{}} 
\toprule
\multirow{2}{*}{\textbf{Variants}} & \multicolumn{3}{c|}{\textbf{SWaT}}                                  & \multicolumn{3}{c|}{\textbf{WADI}}                                                                         & \multicolumn{3}{c|}{\textbf{SMAP}}                                  & \multicolumn{3}{c}{\textbf{MSL}}                                   \\ \cmidrule(l){2-4} \cmidrule(l){5-7} \cmidrule(l){8-10} \cmidrule(l){11-13} 
                                  & \textbf{P(\%)}       & \textbf{R(\%)}       & \textbf{F1(\%)}          & \multicolumn{1}{c}{\textbf{P(\%)}} & \multicolumn{1}{c}{\textbf{R(\%)}} & \multicolumn{1}{c|}{\textbf{F1}} & \textbf{P(\%)}       & \textbf{R(\%)}       & \textbf{F1(\%)}          & \textbf{P(\%)}       & \textbf{R(\%)}       & \textbf{F1(\%)}          \\ \midrule
w/ 2-graph                               & 79.29                &     77.73            &    71.68             & \multicolumn{1}{c}{82.34}          & \multicolumn{1}{c}{84.17}           & \multicolumn{1}{c|}{77.62}        &     85.62                 &     87.33                 &      81.97                & 89.11 &  84.35 & 86.48          \\

w/ irr-hypergraph                               & 93.38                 & 94.24                 & 89.95                 & \multicolumn{1}{c}{92.73}           & \multicolumn{1}{c}{89.64}           & \multicolumn{1}{c|}{87.35}        &    93.12                 &     90.87                 &      91.58                    &                     94.58 &  92.15 & 90.65                            \\

w/o PosEmb                            & \multicolumn{1}{l}{94.21} & \multicolumn{1}{l}{91.53} & \multicolumn{1}{l|}{90.27} &   93.17                                 &                                   91.62 &           \textbf{90.51}                      &      94.78                &     93.37                 &      92.96                  &     94.74 &  93.13 & 91.65 \\

w/ SPEs                           & \multicolumn{1}{l}{\textbf{97.84}} & \multicolumn{1}{l}{96.11} & \multicolumn{1}{l|}{\textbf{93.26}} &           91.85                         &                                   87.91 &         88.31                        &     92.64                 &     94.83                 &      93.57            &      94.83 &  91.65 & 92.58 \\

$\textbf{w/o} \hspace{1mm} \mathbf{z_{i}}$                          & \multicolumn{1}{l}{81.17} & \multicolumn{1}{l}{83.81} & \multicolumn{1}{l|}{80.35} &                80.02                    &    79.71                                &    77.86                             &                84.73                 &     87.61                 &      83.42                 &      82.63            &      85.53 &  82.37                  \\

$\textbf{w/o} \hspace{1mm} \mathbf{\alpha_{p, i}}, \mathbf{\beta_{i, p}, \gamma_{i, p}, \zeta_{i, p}}$                               & 78.16                & 76.37 & 73.94                 & \multicolumn{1}{c}{77.81}           & \multicolumn{1}{c}{75.32}           & \multicolumn{1}{c|}{71.32}        &  82.94                 &     81.78                 &      80.56         &       83.45           &      80.53 &  78.64                 \\

w/ proj                             & 92.32                & 88.43                & 85.56                 & \multicolumn{1}{c}{90.58}          & \multicolumn{1}{c}{89.73}          & \multicolumn{1}{c|}{87.26}        & 89.72                 &     88.63                 &      86.91                  &    93.67           &      91.74 &  88.35  \\

w/o reg                        & 93.24                & 94.17                & 91.65                 & \multicolumn{1}{c}{88.73}          & \multicolumn{1}{c}{89.31}          & \multicolumn{1}{c|}{84.31}        &   93.71                 &     91.54                 &      90.65                &       92.53           &      93.62 &  87.19  \\

w/o $\mathbf{\delta_{i,p}}$                         & 87.97                &  85.66                & 86.15                 & \multicolumn{1}{c}{82.42}          & \multicolumn{1}{c}{85.73}          & \multicolumn{1}{c|}{84.95}        &        87.35                 &     88.73                 &      86.41                  &               84.37           &      86.15 &  83.69               \\

Proposed                          & \multicolumn{1}{l}{\text{97.48}} & \multicolumn{1}{l}{\textbf{98.60}} & \multicolumn{1}{l|}{\text{92.73}} &               \multicolumn{1}{l}{\textbf{98.75}}                     &       \multicolumn{1}{l}{\textbf{92.13}}                             &   \multicolumn{1}{l|}{89.82}                              & \multicolumn{1}{l}{\textbf{98.05}} & \multicolumn{1}{l}{\textbf{99.07}} & \multicolumn{1}{l|}{\textbf{99.23}} & \multicolumn{1}{l}{\textbf{98.26}} & \multicolumn{1}{l}{\textbf{97.11}} & \multicolumn{1}{l}{\textbf{96.84}} \\ \bottomrule
\multicolumn{8}{l}{Best performance in bold} \\ 
\multicolumn{8}{l}{} \\
\end{tabular}
}
\end{table*}

\vspace{-2mm}
(d) We study the importance of attention-mechanism in computing the discriminative representations($x^{(t)}_{v_{i}}$) of the hypergraph-structured data. We deactivate the attention-mechanism of the HgCNN operator(refer to Equations \ref{eq:hgcnn1} and \ref{eq:hgcnn2}), the HgPool operator(Equation \ref{eq:hgpool5}), and the UnHgPool operator(Equation \ref{eq:unhgpool1}). We perform neighborhood aggregation through the unweighted sum-pooling operation. We refer to the $\textbf{HgAD}$ model in the absence of the attention-mechanism to compute the hypergraph representations as follows, 

\begin{itemize}
\item $\textbf{w/o} \hspace{1mm} \mathbf{\alpha_{p, i}}, \mathbf{\beta_{i, p}, \zeta_{i, p}, \gamma_{i, p}}$: \hspace{-1mm}$\textbf{HgAD}$ model without the attention-mechanism.
\end{itemize}

Eliminating the attention-mechanism results in poor performance of the variant across all the datasets, as reflected in Table \ref{tab:ablation}. In particular, it decreases the precision-score  w.r.t. the $\textbf{HgAD}$ model by more than $19.8\%$ on SWaT; $17.8\%$ on WADI. The hypergraph-attention mechanism performs weighted relational learning on the sensor-topology while taking into account the importance of the hypernodes and hyperedges. As a result, it leads to better generalization and performance on the hypergraph-structured data. (e) We investigate the effectiveness of the importance measure(refer to Equation \ref{eq:hgpool2}) and the impact of the regularization technique(refer to Equation \ref{eq:hgpool4}) in the HgPool operator. We refer to the $\textbf{HgAD}$ models to compute the importance score through the projection vector method\cite{gao2019graph} and with disabled regularization technique in the Hgpool operator are as follows: 

\vspace{-1mm}
\begin{itemize} 
\item $\textbf{w/} \hspace{1mm}  \textbf{proj}$: $\textbf{HgAD}$ model with learnable projection vector.
\item $\textbf{w/o} \hspace{1mm} \textbf{reg}$: $\textbf{HgAD}$ model without the regularization.
\end{itemize} 

\vspace{-1mm}
As can be seen in Table \ref{tab:ablation}, the results indicate the on-par performance of $\textbf{w/} \hspace{1mm} \textbf{proj}$ in precision-score compared to the $\textbf{HgAD}$ model. In the absence of a regularization technique, the variant $\textbf{w/o} \hspace{1mm} \textbf{reg}$ yields an overall relatively $4.3\%$ on SWaT; $6.4\%$ on WADI lower precision score w.r.t. the $\textbf{HgAD}$ model. The local-hypergraph pooling mechanism and a regularization technique help in learning the robust representations of hypergraphs by overcoming the limitations of the low-pass property of HgCNNs. (f)  We analyze the usefulness of attention-weights($\delta_{i, p}$) to model the sooner or later effects of anomaly traversal among the network of interconnected sensors. Please revisit Equation \ref{eq:unhgpool3} in subsection\ref{unhgpool}. We refer to the $\textbf{HgAD}$ model to compute the refined hypernode representations through the UnHgpool operator without a mechanism to capture the anomaly propagation delay or advancement in $\delta_{i, p}$ as follows: 

\vspace{-1mm}
\begin{itemize}
\item $\textbf{w/} \hspace{1mm} \mathbf{\delta_{i, p}}$: \hspace{-1mm}$\textbf{HgAD}$ model with $\delta_{i, p} = \big(\mathbf{W}_{12}\mathbf{g}_{i}^{(t)}\big)^{\top} \mathbf{W}_{12}\mathbf{g}_{p}^{(t)}$. 
\end{itemize}

\vspace{-1mm}
The degradation of the variant performance is evident in Table \ref{tab:ablation}, which shows a drop of $9.7\%$ on SWaT; $13.1\%$ on WADI on precision-score w.r.t. the $\textbf{HgAD}$ model. These verify the effectiveness of our proposed method to capture anomaly propagation delay or advancement. The ablation studies show that the modules of our framework are domain-agnostic and contribute significantly to improved performance of the model. 

\vspace{-4mm}
\subsection{Root-cause analysis}
\vspace{-2mm}
\label{sec:laomalies} 
The traditional data-driven modeling approaches lack explicit contextual information to identify the cause of anomalies obtained from several interdependent systems. Here, we present our approach for uncovering the prominent substructures underlying the hypergraph-structured data for anomaly diagnosis underneath the model predictions. The hypernodes define computation hypergraphs based on their local network neighborhoods inferred from the learned hypergraph representation of the complex sensor networks. The computation hypergraphs obtained from the subhypergraphs explain the root-cause analysis. The abnormality events detected at the hypernodes will traverse along their respective computational hypergraphs via the top-bottom approach. The anomaly information propagates from the root hypernode(deviating sensor) through the intermediate hypernodes all the way to the leaf hypernodes(k-hop or k-order proximities).  k is pre-determined by the user or fixed by the algorithm. In summary, the module presents a bifold approach. It identifies the hypernode deviating from the learned normality relationship with a high anomaly score. Highlights the salient regions of substructures through the computational hypergraphs to diagnose the anomalies underlying the abnormal system.  

\vspace{-3mm}
\subsection{Prescriptive analytics}
\label{sec:interanomalies} 
\vspace{-2mm}
Anomalies have cascading effects on the overall performance of large-scale systems. We present an offline, \textbf{hypergraph-based predictive control(HgPC)} module to learn the optimal control policy through solving a single-objective optimization task to offset an anomaly. The module predicts the sequence of the manipulated variable, i.e., hypernode(deviating sensor) values so that the proposed framework predicts normal behavior of the system or falls-inline with the expected behavior. We leverage genetic algorithms(GAs) to recommend optimal actions.

\vspace{-3mm}
\begin{figure}[h]
    \centering
    \includegraphics[scale=0.5]{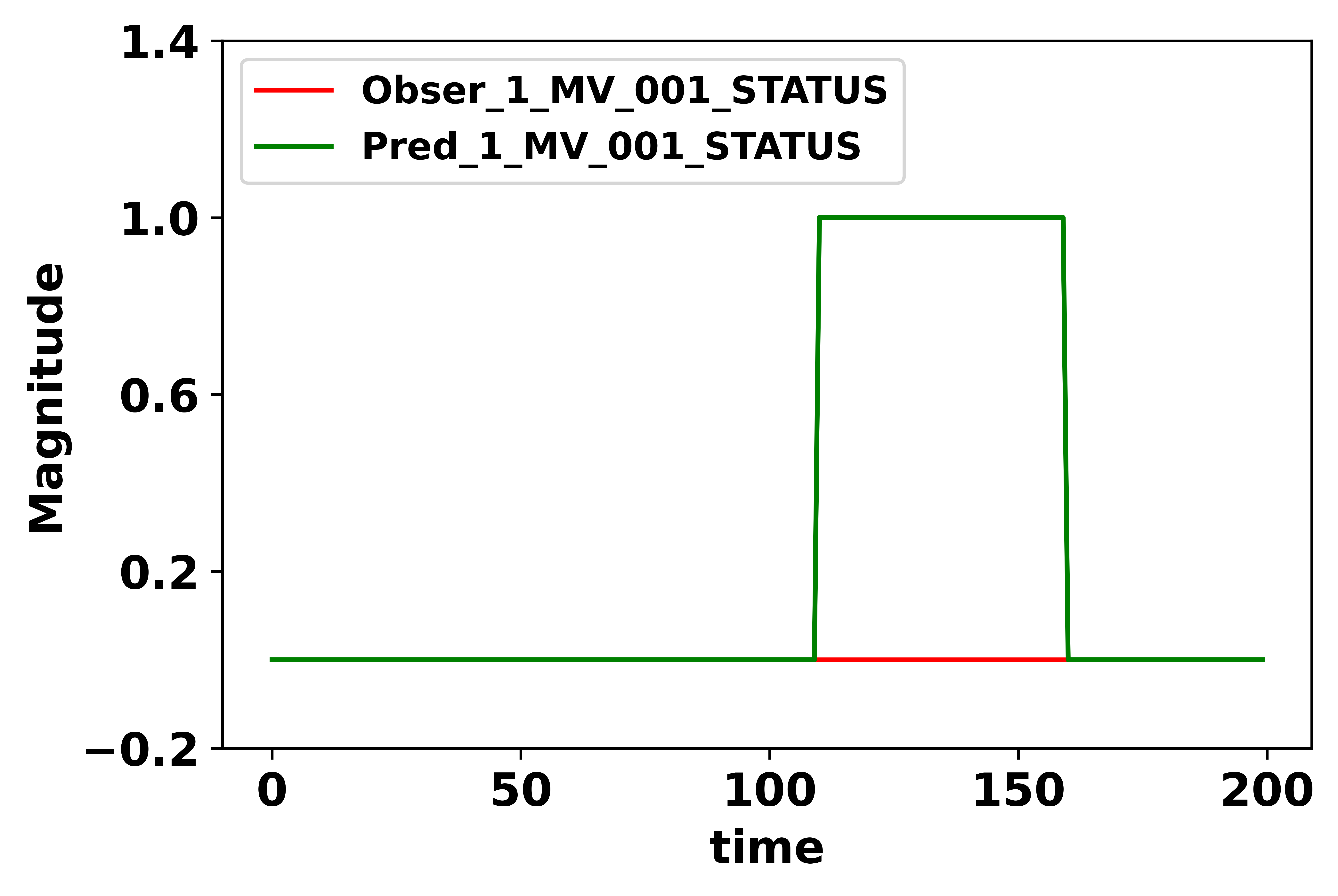}
    \centering
    \includegraphics[scale=0.5]{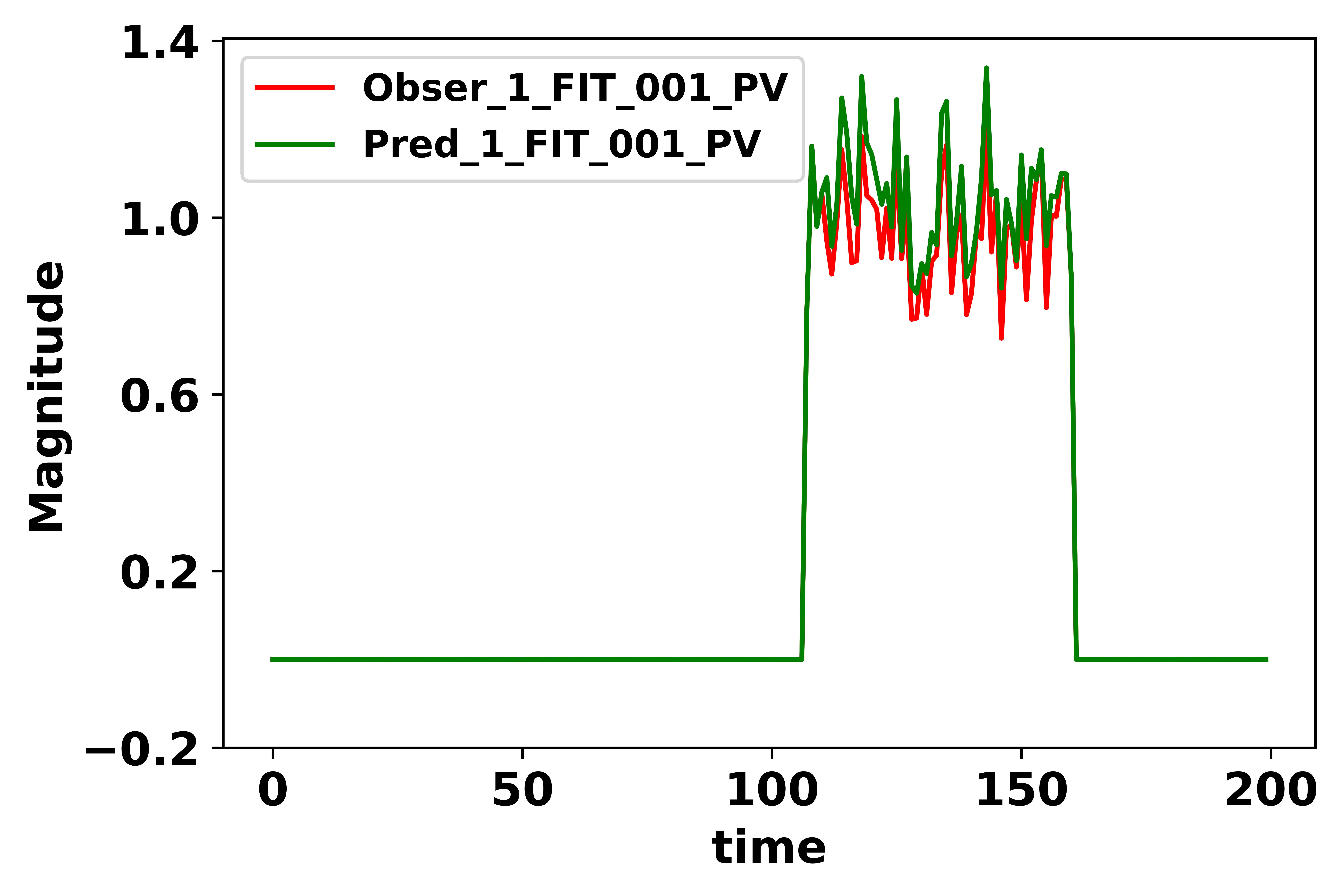}
    \vspace{-1mm}
    \caption{Comparison of model predictions and the ground-truth}
    \label{fig:rc-cs}
\end{figure}

\vspace{-7mm}
\begin{figure}[h]
    \centering
    \includegraphics[scale=0.45]{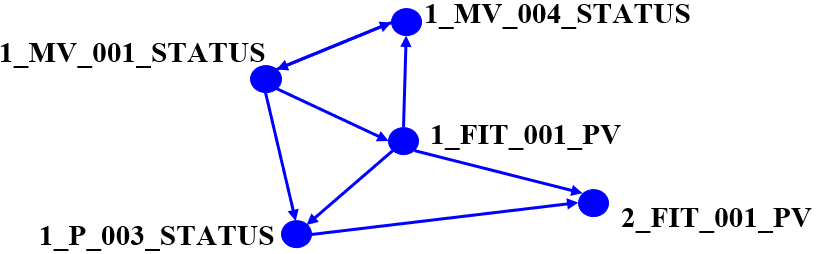}
    \vspace{-1mm}
    \caption{A subhypergraph of the sensor network.}
    \label{fig:chg}
\end{figure}

\vspace{8mm}
\subsection{Case study} 
\vspace{-2mm}
We conduct a case study involving an anomaly with a known deviation behavior. A subunit of WADI consists of a flow-rate manipulator \texttt{1\_MV\_001\_STATUS} and a flow indicator transmitter \texttt{1\_FIT\_001\_PV}. Consider the first-anomaly scenario(refer to documentation\footnote{{https://itrust.sutd.edu.sg/itrust-labs\_datasets/}}) for the \texttt{1\_MV\_001\_STATUS}. The HgED module predicted an increase as \texttt{1\_FIT\_001\_PV} shoot-ups. The learned-normality relationship suggests that the variables are positively correlated. Due to the adversarial attack, no abrupt change was observed in \texttt{1\_MV\_001\_STATUS} and thus leading to a high forecast error. We predict based on the high anomalous score \texttt{1\_MV\_001\_STATUS} as the deviation-sensor. From the computation hypergraph, we infer that the \texttt{1\_FIT\_001\_PV} was under malicious attack. The Figure\ref{fig:rc-cs} shows the model sensor forecasts. The process-plant operations advisor, the HgPC module, suggests reducing the  \texttt{1\_MV\_001\_STATUS} to 0 to offset the anomaly. Figure\ref{fig:chg} shows the subhypergraph to explain the root-cause analysis.  

\vspace{-3mm}
\section{Conclusion}
\label{sec:conclusion}   
\vspace{-3mm}
We present a scalable, domain-agnostic hypergraph framework for anomaly detection on generic multisensor data. The framework incentivizes joint learning of the optimal hypergraph-structure representation, temporal trends, and spatial dependencies in the sensor networks for distinguishing the normal-abnormal instances. We also present the root cause analysis and suggest recommendations for better insights and decisions to remedy the anomalies. The experimental results support our framework efficacy to achieve better performance on the high-dimensional time-series anomaly-detection task.

\vspace{-3mm}
{
\small
\bibliographystyle{plain}
\bibliography{sample-base.bib}

\begin{thebibliography}{10}

\bibitem{aggarwal2017introduction}
Charu~C Aggarwal.
\newblock An introduction to outlier analysis.
\newblock In {\em Outlier analysis}, pages 1--34. Springer, 2017.

\bibitem{anderson1976time}
OD~Anderson.
\newblock Time-series. 2nd edn., 1976.

\bibitem{angiulli2002fast}
Fabrizio Angiulli and Clara Pizzuti.
\newblock Fast outlier detection in high dimensional spaces.
\newblock In {\em European conference on principles of data mining and
  knowledge discovery}, pages 15--27. Springer, 2002.

\bibitem{bai2018empirical}
Shaojie Bai, J~Zico Kolter, and Vladlen Koltun.
\newblock An empirical evaluation of generic convolutional and recurrent
  networks for sequence modeling.
\newblock {\em arXiv preprint arXiv:1803.01271}, 2018.

\bibitem{breunig2000lof}
Markus~M Breunig, Hans-Peter Kriegel, Raymond~T Ng, and J{\"o}rg Sander.
\newblock Lof: identifying density-based local outliers.
\newblock In {\em Proceedings of the 2000 ACM SIGMOD international conference
  on Management of data}, pages 93--104, 2000.

\bibitem{chen2021learning}
Zekai Chen, Dingshuo Chen, Xiao Zhang, Zixuan Yuan, and Xiuzhen Cheng.
\newblock Learning graph structures with transformer for multivariate time
  series anomaly detection in iot.
\newblock {\em IEEE Internet of Things Journal}, 2021.

\bibitem{deng2021graph}
Ailin Deng and Bryan Hooi.
\newblock Graph neural network-based anomaly detection in multivariate time
  series.
\newblock In {\em Proceedings of the AAAI Conference on Artificial
  Intelligence}, volume~35, pages 4027--4035, 2021.

\bibitem{feng2019hypergraph}
Yifan Feng, Haoxuan You, Zizhao Zhang, Rongrong Ji, and Yue Gao.
\newblock Hypergraph neural networks.
\newblock In {\em Proceedings of the AAAI conference on artificial
  intelligence}, volume~33, pages 3558--3565, 2019.

\bibitem{fu2022mad}
Yiwei Fu and Feng Xue.
\newblock Mad: Self-supervised masked anomaly detection task for multivariate
  time series.
\newblock {\em arXiv preprint arXiv:2205.02100}, 2022.

\bibitem{gao2019graph}
Hongyang Gao and Shuiwang Ji.
\newblock Graph u-nets.
\newblock In {\em international conference on machine learning}, pages
  2083--2092. PMLR, 2019.

\bibitem{hochreiter1997long}
Sepp Hochreiter and J{\"u}rgen Schmidhuber.
\newblock Long short-term memory.
\newblock {\em Neural computation}, 9(8):1735--1780, 1997.

\bibitem{hundman2018detecting}
Kyle Hundman, Valentino Constantinou, Christopher Laporte, Ian Colwell, and Tom
  Soderstrom.
\newblock Detecting spacecraft anomalies using lstms and nonparametric dynamic
  thresholding.
\newblock In {\em Proceedings of the 24th ACM SIGKDD international conference
  on knowledge discovery \& data mining}, pages 387--395, 2018.

\bibitem{lazarevic2005feature}
Aleksandar Lazarevic and Vipin Kumar.
\newblock Feature bagging for outlier detection.
\newblock In {\em Proceedings of the eleventh ACM SIGKDD international
  conference on Knowledge discovery in data mining}, pages 157--166, 2005.

\bibitem{lee2021fnet}
James Lee-Thorp, Joshua Ainslie, Ilya Eckstein, and Santiago Ontanon.
\newblock Fnet: Mixing tokens with fourier transforms.
\newblock {\em arXiv preprint arXiv:2105.03824}, 2021.

\bibitem{li2018anomaly}
Dan Li, Dacheng Chen, Jonathan Goh, and See-kiong Ng.
\newblock Anomaly detection with generative adversarial networks for
  multivariate time series.
\newblock {\em arXiv preprint arXiv:1809.04758}, 2018.

\bibitem{li2019mad}
Dan Li, Dacheng Chen, Baihong Jin, Lei Shi, Jonathan Goh, and See-Kiong Ng.
\newblock Mad-gan: Multivariate anomaly detection for time series data with
  generative adversarial networks.
\newblock In {\em International conference on artificial neural networks},
  pages 703--716. Springer, 2019.

\bibitem{li2019enhancing}
Shiyang Li, Xiaoyong Jin, Yao Xuan, Xiyou Zhou, Wenhu Chen, Yu-Xiang Wang, and
  Xifeng Yan.
\newblock Enhancing the locality and breaking the memory bottleneck of
  transformer on time series forecasting.
\newblock {\em Advances in neural information processing systems}, 32, 2019.

\bibitem{li2021multivariate}
Zhihan Li, Youjian Zhao, Jiaqi Han, Ya~Su, Rui Jiao, Xidao Wen, and Dan Pei.
\newblock Multivariate time series anomaly detection and interpretation using
  hierarchical inter-metric and temporal embedding.
\newblock In {\em Proceedings of the 27th ACM SIGKDD Conference on Knowledge
  Discovery \& Data Mining}, pages 3220--3230, 2021.

\bibitem{liu2008isolation}
Fei~Tony Liu, Kai~Ming Ting, and Zhi-Hua Zhou.
\newblock Isolation forest.
\newblock In {\em 2008 eighth ieee international conference on data mining},
  pages 413--422. IEEE, 2008.

\bibitem{mirsky2018kitsune}
Yisroel Mirsky, Tomer Doitshman, Yuval Elovici, and Asaf Shabtai.
\newblock Kitsune: an ensemble of autoencoders for online network intrusion
  detection.
\newblock {\em arXiv preprint arXiv:1802.09089}, 2018.

\bibitem{park2018multimodal}
Daehyung Park, Yuuna Hoshi, and Charles~C Kemp.
\newblock A multimodal anomaly detector for robot-assisted feeding using an
  lstm-based variational autoencoder.
\newblock {\em IEEE Robotics and Automation Letters}, 3(3):1544--1551, 2018.

\bibitem{ruff2018deep}
Lukas Ruff, Robert Vandermeulen, Nico Goernitz, Lucas Deecke, Shoaib~Ahmed
  Siddiqui, Alexander Binder, Emmanuel M{\"u}ller, and Marius Kloft.
\newblock Deep one-class classification.
\newblock In {\em International conference on machine learning}, pages
  4393--4402. PMLR, 2018.

\bibitem{scholkopf2001estimating}
Bernhard Sch{\"o}lkopf, John~C Platt, John Shawe-Taylor, Alex~J Smola, and
  Robert~C Williamson.
\newblock Estimating the support of a high-dimensional distribution.
\newblock {\em Neural computation}, 13(7):1443--1471, 2001.

\bibitem{shen2020timeseries}
Lifeng Shen, Zhuocong Li, and James Kwok.
\newblock Timeseries anomaly detection using temporal hierarchical one-class
  network.
\newblock {\em Advances in Neural Information Processing Systems},
  33:13016--13026, 2020.

\bibitem{shin2020itad}
Youjin Shin, Sangyup Lee, Shahroz Tariq, Myeong~Shin Lee, Okchul Jung, Daewon
  Chung, and Simon~S Woo.
\newblock Itad: integrative tensor-based anomaly detection system for reducing
  false positives of satellite systems.
\newblock In {\em Proceedings of the 29th ACM international conference on
  information \& knowledge management}, pages 2733--2740, 2020.

\bibitem{shyu2003novel}
Mei-Ling Shyu, Shu-Ching Chen, Kanoksri Sarinnapakorn, and LiWu Chang.
\newblock A novel anomaly detection scheme based on principal component
  classifier.
\newblock Technical report, Miami Univ Coral Gables Fl Dept of Electrical and
  Computer Engineering, 2003.

\bibitem{su2019robust}
Ya~Su, Youjian Zhao, Chenhao Niu, Rong Liu, Wei Sun, and Dan Pei.
\newblock Robust anomaly detection for multivariate time series through
  stochastic recurrent neural network.
\newblock In {\em Proceedings of the 25th ACM SIGKDD international conference
  on knowledge discovery \& data mining}, pages 2828--2837, 2019.

\bibitem{tariq2019detecting}
Shahroz Tariq, Sangyup Lee, Youjin Shin, Myeong~Shin Lee, Okchul Jung, Daewon
  Chung, and Simon~S Woo.
\newblock Detecting anomalies in space using multivariate convolutional lstm
  with mixtures of probabilistic pca.
\newblock In {\em Proceedings of the 25th ACM SIGKDD international conference
  on knowledge discovery \& data mining}, pages 2123--2133, 2019.

\bibitem{vaswani2017attention}
Ashish Vaswani, Noam Shazeer, Niki Parmar, Jakob Uszkoreit, Llion Jones,
  Aidan~N Gomez, {\L}ukasz Kaiser, and Illia Polosukhin.
\newblock Attention is all you need.
\newblock {\em Advances in neural information processing systems}, 30, 2017.

\bibitem{xu2021anomaly}
Jiehui Xu, Haixu Wu, Jianmin Wang, and Mingsheng Long.
\newblock Anomaly transformer: Time series anomaly detection with association
  discrepancy.
\newblock {\em arXiv preprint arXiv:2110.02642}, 2021.

\bibitem{yadati2019hypergcn}
Naganand Yadati, Madhav Nimishakavi, Prateek Yadav, Vikram Nitin, Anand Louis,
  and Partha Talukdar.
\newblock Hypergcn: A new method for training graph convolutional networks on
  hypergraphs.
\newblock {\em Advances in neural information processing systems}, 32, 2019.

\bibitem{yairi2017data}
Takehisa Yairi, Naoya Takeishi, Tetsuo Oda, Yuta Nakajima, Naoki Nishimura, and
  Noboru Takata.
\newblock A data-driven health monitoring method for satellite housekeeping
  data based on probabilistic clustering and dimensionality reduction.
\newblock {\em IEEE Transactions on Aerospace and Electronic Systems},
  53(3):1384--1401, 2017.

\bibitem{zhanggrelen}
Weiqi Zhang, Chen Zhang, and Fugee Tsung.
\newblock Grelen: Multivariate time series anomaly detection from the
  perspective of graph relational learning.
\newblock {\em https://www.ijcai.org/proceedings/2022/0332.pdf}, 2022.

\bibitem{zhao2020multivariate}
Hang Zhao, Yujing Wang, Juanyong Duan, Congrui Huang, Defu Cao, Yunhai Tong,
  Bixiong Xu, Jing Bai, Jie Tong, and Qi~Zhang.
\newblock Multivariate time-series anomaly detection via graph attention
  network.
\newblock In {\em 2020 IEEE International Conference on Data Mining (ICDM)},
  pages 841--850. IEEE, 2020.

\bibitem{zhou2019beatgan}
Bin Zhou, Shenghua Liu, Bryan Hooi, Xueqi Cheng, and Jing Ye.
\newblock Beatgan: Anomalous rhythm detection using adversarially generated
  time series.
\newblock In {\em IJCAI}, pages 4433--4439, 2019.

\bibitem{zong2018deep}
Bo~Zong, Qi~Song, Martin~Renqiang Min, Wei Cheng, Cristian Lumezanu, Daeki Cho,
  and Haifeng Chen.
\newblock Deep autoencoding gaussian mixture model for unsupervised anomaly
  detection.
\newblock In {\em International conference on learning representations}, 2018.

\end{thebibliography}
}


\end{document}